\documentclass{article} 

\usepackage[accepted]{icml2020}
\usepackage{times}  
\usepackage{helvet} 
\usepackage{courier}  
\usepackage[hyphens]{url}  
\usepackage{graphicx} 
\usepackage{graphicx}  

\usepackage{hyperref}       
\usepackage{url}            
\usepackage{booktabs}       
\usepackage{nicefrac}       
\usepackage{microtype}      
\usepackage{tikz}

\usepackage{wrapfig}

\usepackage{times}
\usepackage{xcolor}
\usepackage{soul}

\usepackage{graphicx}
\usepackage{amsmath}
\usepackage{amsfonts,amssymb}
\usepackage{multirow}

\newcommand{\ignore}[1]{{}}
\newcommand{\hankz}[1]{{\color{blue} #1 \color{black}}}
\newcommand{\fengwf}[1]{{\color{magenta} #1 \color{black}}}
\newcommand{\grey}[1]{{\textcolor[rgb]{0.55,0.55,0.55}{#1}}}

\def\ours{{\tt FedRL}}

\icmltitlerunning{Federated Collaborative Reinforcement Learning}
\makeatletter
 \def\@textbottom{\vskip \z@ \@plus 1pt}
 \let\@texttop\relax
\makeatother
\begin{document}

\twocolumn[
\icmltitle{Federated Deep Reinforcement Learning}




\begin{icmlauthorlist}
\icmlauthor{Hankz Hankui Zhuo}{to}
\icmlauthor{Wenfeng Feng}{to}
\icmlauthor{Yufeng Lin}{to}
\icmlauthor{Qian Xu}{goo}
\icmlauthor{Qiang Yang}{goo}
\end{icmlauthorlist}

\icmlaffiliation{to}{zhuohank@mail.sysu.edu.cn, fengwf2014@outlook.com, linyf23@mail2.sysu.edu.cn, Sun Yat-Sen University, Guangzhou, China;}
\icmlaffiliation{goo}{\{qianxu,qiangyang\}@webank.com, WeBank, Shenzhen, China}

\icmlcorrespondingauthor{Hankz Hankui Zhuo}{zhuohank@mail.sysu.edu.cn}


\vskip 0.3in
]
\printAffiliationsAndNotice



\begin{abstract}
In deep reinforcement learning, building policies of high-quality is challenging when the feature space of states is small and the training data is limited. Despite the success of previous transfer learning approaches in deep reinforcement learning, directly transferring data or models from an agent to another agent is often not allowed due to the privacy of data and/or models in many privacy-aware applications. In this paper, we propose a novel deep reinforcement learning framework to federatively build models of high-quality for agents with consideration of their privacies, namely Federated deep Reinforcement Learning ({\ours}). To protect the privacy of data and models, we exploit Gausian differentials on the information shared with each other when updating their local models. In the experiment, we evaluate our {\ours} framework in two diverse domains, Grid-world and Text2Action domains, by comparing to various baselines.
\end{abstract}

\section{Introduction}
In deep reinforcement learning, building policies of high-quality is challenging when the feature space of states is small and the training data is limited. In many real-world applications, however, datasets from clients are often privacy sensitive \cite{DBLP:conf/nips/DuchiJW12} and it is often difficult for such a data center to guarantee building models of high-quality. To deal with the issue, Konecny et al. propose a new learning setting, namely federated learning, whose goal is to train a classification or clustering model with training data involving texts, images or videos distributed over a large number of clients \cite{DBLP:journals/corr/KonecnyMYRSB16,DBLP:conf/aistats/McMahanMRHA17}. Different from previous federated learning setting (c.f. \cite{DBLP:journals/tist/YangLCT19}), we propose a novel federated learning framework based on reinforcement learning \cite{DBLP:books/lib/SuttonB98,Mnih2015Human,pmlr-v80-co-reyes18a}, i.e., \emph{\textbf{Fed}erated deep \textbf{R}einforcement \textbf{L}earning} ({\ours}), which aims to learn a private Q-network policy for each agent by sharing limited information (i.e., output of the Q-network) among agents. The information is ``encoded'' when it is sent to others and ``decoded'' when it is received by others. We assume that some agents have \emph{rewards} corresponding to states and actions, while others have only observed states without \emph{rewards}. Without rewards, those agents are unable to build decision policies on their own information. We claim that all agents benefit from joining the federation in building decision policies. 

There are many applications regarding federated reinforcement learning. \emph{For example, in the manufacturing industry, producing products may involve various factories which produce different components of the products. Factories' decision policies are private and will not be shared with each other. On the other hand, building individual decision policies of high-quality on their own is often difficult due to their limited businesses and lack of rewards (for some factories). It is thus helpful for them to learn decision polices federatively under the condition that private data is not given away. Another example is building medical treatment policies to patients for hospitals. Patients may be treated in some hospitals and never give feedbacks to the treatments, which indicates these hospitals are unable to collect rewards based on the treatments given to the patients and build treatment decision policies for patients. In addition, data records about patients are private and may not be shared among hospitals. It is thus necessitated to learn treatment policies for hospitals federatively. 
}

Our {\ours} framework is different from multi-agent reinforcement learning, which is concerned with a set of autonomous agents that observe global states (or partial states which are directly shared to make ``global'' states), select an individual action and receive a team reward (or each agent receives an individual reward but shares it with other agents) \cite{DBLP:journals/corr/TampuuMKKKAAV15,DBLP:conf/atal/LeiboZLMG17,DBLP:conf/nips/FoersterAFW16}.  {\ours} assumes agents do not share their partial observations and some agents are unable to receive rewards. Our {\ours} framework is also different from transfer learning in reinforcement learning, which aims to transfer experience gained in learning to perform one task to help improve learning performance in a related but different task or agent, assuming observations are shared with each other \cite{Taylor:2009,NIPS2018_7856}, while {\ours} assumes states cannot be shared among agents.

Our {\ours} framework functions in three phases. Initially, each agent collects output values of Q-networks from other agents, which are ``encrypted'' with Gausian differentials. Furthermore, it builds a shared value network, e.g., MLP (multilayer perceptron), to compute a global Q-network output with its own Q-network output and the encrypted values as input. Finally, it updates both the shared value network and its own Q-network based on the global Q-network output. Note that MLP is shared among agents while agents' own Q-networks are unknown to others and should not be inferred based on the encrypted Q-network output shared in the training process. 

In the remainder of the paper, we first review previous work related to our {\ours} framework, and then present the problem formulation of {\ours}. After that, we introduce our {\ours} framework in detail. Finally we evaluate our {\ours} framework in the Grid-World domain with various sizes and the Text2Actions domain.

\ignore{
Deep reinforcement learning aims to learn complex skills or decision policies from observations \cite{DBLP:journals/corr/MnihBMGLHSK16,DBLP:journals/jmlr/LevineFDA16}. Despite the success of previous approaches, however, it is often difficult or time-consuming to learn an individual policy of high-quality due to its limited observed data (numbers of features and instances are both limited) in many complex domains involving multiple agents.

A higher-level policy that is provided with temporally extended and intelligent behaviours can reason at a higher level of abstraction and solve more temporally-extended tasks. Furthermore, the same lower-level skills could be reused to accomplish multiple tasks efficiently.

\cite{ijcai2018-774}

\cite{DBLP:conf/aaai/Bou-AmmarERT15} 

Deep reinforcement learning has been demonstrated effective in many applications, such as ...

It is challenging when there are limited training data or exploration-exploitation experiences from environments. There have been many approaches to learn policies from limited training data, such as transfer reinforcement learning \cite{}, multi-agent reinforcement learning \cite{}

distributed stochastic gradient descent \cite{}, distributed multi-task reinforcement learning \cite{}. 

Current solutions cannot handle the issues of model or data privacies. 

We thus need to build a federation of agents to do reinforcement learning.

Present an example to motivate federated reinforcement learning: medical diagnosis.

Present our solutions.
}

\section{Related Work}
The nascent field of federated learning considers training statistical models directly on devices \cite{DBLP:journals/corr/KonecnyMR15,DBLP:conf/aistats/McMahanMRHA17}. The aim in federated learning is to fit a model to data generated by distributed nodes. Each node collects data in a non-IID manner across the network, with data on each node being generated by a distinct distribution. There are typically a large number of nodes in the network, and communication is often a significant bottleneck. Different from previous work that train a single global model across the network \cite{DBLP:journals/corr/KonecnyMR15,DBLP:journals/corr/KonecnyMYRSB16,DBLP:conf/aistats/McMahanMRHA17}, Smith et al. propose to learn separate models for each node which is naturally captured through a multi-task learning (MTL) framework, where the goal is to consider fitting separate but related models simultaneously \cite{DBLP:conf/nips/SmithCST17}. Different from those federated learning approaches, we consider federated settings in reinforcement learning.

Our work is also related to multi-agent reinforcement learning (MARL) which involves a set of agents in a shared environment. A straightforward way to MARL is to extend the single-agent RL approaches. Q-learning has been extended to cooperative multi-agent settings, namely Independent Q-learning (IQL), in which each agent observes the global state, selects an individual action and receives a team reward \cite{DBLP:journals/corr/TampuuMKKKAAV15,DBLP:conf/atal/LeiboZLMG17}. One challenging of MARL is that multi-agent domains are non-stationary from agent's perspectives, due to other agents' interactions in the shared environment. To address this issue, \cite{DBLP:conf/icml/OmidshafieiPAHV17} propose to explore Concurrent Experience Replay Trajectories (CERTs) structures, which store different agents' histories, and align them together based on the episode indices and time steps. Due to the action space growing exponentially with the number of agents, learning becomes very difficult due to partial observability of limited communication when the number of agents is large. \cite{DBLP:conf/nips/LoweWTHAM17} thus propose to solve the MARL problem through a Centralized Critic and a Decentralized Actor, and \cite{DBLP:conf/icml/RashidSWFFW18} propose to exploit a linear decomposition of the joint value function across agents. Different from MARL, our {\ours} framework assumes agents do not share their partial observations and some agents are unable to receive rewards, instead of assuming observations are sharable and all agents are able to receive rewards.

\ignore{
\cite{pmlr-v80-co-reyes18a}

Reinforcement learning (RL) together with supervised learning and unsupervised learning are the fundamental branches of machine learning. RL problems can be typically formalized by Markov Decision Processes (MDPs) in which the agent continuously interact with the environment and receive some feedback (reward) from the environment. The agent's task is to maximize the cumulative reward (the total reward it obtains in the long run). The model of the MDPs, which describes the dynamics of the environment, is usually unknown, so the methods that solve the MDPs are called model-free methods. DQN \cite{Mnih2015Human}, which adopts a convolutional neural network to approximate the Q-function, is one of the most popular model-free method. There are many extensions of DQN, such as DRQN \cite{DBLP:conf/aaaifs/HausknechtS15} which make use of recurrent neural networks when approximating the Q-function. There are another line of model-free approaches that consider not only the value function but also the explicit policy. They are called Actor-Critic \cite{DBLP:books/lib/SuttonB98} because they consider the value function as a critic and the policy as an actor. Recently, there are some varieties, e.g. A3C \cite{DBLP:conf/icml/MnihBMGLHSK16} which considers the advantageous action-value function.}

\section{Problem Definition}
A Markov Decision Process (MDP) can be defined by $\langle S, A, T, r\rangle$, where $S$ is the state space, and $A$ is the action space. $T$ is the transition function: $S \times A \rightarrow S$, i.e., $T(s,a,s') = P(s'|s,a)$, specifying the probability of next state $s' \in S$ given current state $s \in S$ and $a \in A$ that applies on $s$. $r$ is the reward function: $S \rightarrow \mathcal{R}$, where $\mathcal{R}$ is the space of real numbers. 
Given a policy $\pi:S\rightarrow A$, the value function $V^{\pi}(s)$ and the Q-function $Q^{\pi}(s,a)$ at step $t+1$, can be updated by their $t$ step: \[V_{t+1}^{\pi}(s)=r(s)+\sum_{s'\in S}T(s,\pi(s),s')V_t^{\pi}(s'),\] 
and 
\[Q_{t+1}^{\pi}(s,a)=r(s)+\sum_{s'\in S}T(s,a,s')V_t^{\pi}(s'),\] 
for $t\in\{0,\ldots,K-1\}$. The solution to an MDP problem is the best policy $\pi^*$ such that $V^{\pi^*}(s)=\max_{\pi}V^{\pi}(s)$ or $Q^{\pi^*}(s,\pi^*(s))=\max_{\pi}Q^{\pi}(s,\pi(s))$. In DQN (Deep Q-Network), given transition function $T$ is unknown, the Q-function is represented by a Q-network $Q(s,a; \theta)$ with $\theta$ as parameters of the network, and updated by \[Q_{t+1}(s,a;\theta)=E_{s'}\Big\{r(s)+\gamma\max_{a'\in A}Q_t(s',a';\theta)|s,a\Big\},\] as done by \cite{Mnih2015Human}. To learn the parameters $\theta$, one way is to store transitions $\langle s,a,s',r\rangle$ in replay memories $\Omega$ and exploit a mini-batch sampling to repeatedly update $\theta$ \cite{Mnih2015Human}. Once $\theta$ is learnt, the policy $\pi^*$ can be extracted from $Q(s,a;\theta)$,
\[\pi^*(s)=\arg\max_{a\in A}Q(s,a;\theta).\]

\ignore{
Suppose there are two agents $\alpha$ and $\beta$ making decisions in two different environments, e.g., two doctors treating patients at two different hospitals, respectively. Agent $\alpha$ is able to collect reward $r$ and new state $s'$ from the environment given current state $s$ and action $a$, i.e., agent $\alpha$ is capable of building complete transitions $\langle s,a,s',r\rangle$ from the environment (e.g., patients are treated in the hospital). On the other hand, agent $\beta$ can only collect current state $s$ and action $a$, without any information about next state $s'$ and reward $r$ from the environment. For example, patients that doctor $\beta$ treats never come back such that $\beta$ cannot collect new state $s'$ about the patients and reward $r$ based on the treatment (i.e., action $a$) $\beta$ gave to the patient (i.e., current state $s$). }

We define our \emph{federated} deep reinforcement learning problem by: \emph{given transitions $\mathcal{D}_{\alpha} = \{\langle s_{\alpha},a_{\alpha},s'_{\alpha},r_{\alpha}\rangle\}$ collected by agent $\alpha$, and pairs of states and actions $\mathcal{D}_{\beta}=\{\langle s_{\beta},a_{\beta}\rangle\}$ collected by agent $\beta$, we aim to federatively build policies $\pi^*_{\alpha}$ and $\pi^*_{\beta}$ for agents $\alpha$ and $\beta$, respectively. } Note that in this paper we consider the federation with two members for simplicity. The setting can be extended to many agents by exploiting the same federated mechanism between each two agents. We denote states, actions, Q-functions, and policies with respect to agents $\alpha$ and $\beta$, by ``$s_{\alpha}\in S_{\alpha}$, $a_{\alpha}\in A_{\alpha}$, $Q_{\alpha}$, $\pi^*_{\alpha}$'' and ``$s_{\beta}\in S_{\beta}$, $a_{\beta}\in A_{\beta}$, $Q_{\beta}$, $\pi^*_{\beta}$'', respectively.  

In our federated deep reinforcement learning problem, we assume:

\textbf{A1:}
The feature spaces of states $s_{\alpha}$ and $s_{\beta}$ are \emph{different} between agents $\alpha$ and $\beta$. For example, a state $s_{\alpha}$ denotes a patient's cardiogram in hospital $\alpha$, while another state $s_{\beta}$ denotes the same patient's electroencephalogram in hospital $\beta$, indicating the feature spaces of $s_{\alpha}$ and $s_\beta$ are different. 

\textbf{A2:} Transitions $\mathcal{D}_\alpha$ and $\mathcal{D}_\beta$ cannot be shared directly between $\alpha$ and $\beta$ when they learning their own models. The \emph{correspondences} between transitions from $\mathcal{D}_\alpha$ and $\mathcal{D}_\beta$ are, however, known to each other. In other words, agent $\alpha$ can send the ``ID'' of a transition to agent $\beta$, and agent $\beta$ can use that ``ID'' to find its corresponding transition in $\mathcal{D}_\beta$. For example, in hospital, ``ID'' can correspond to a specific patient.

\ignore{ states $s_{\alpha}$ and rewards $r_{\alpha}$ cannot be shared with agent $\beta$, and states $s_{\beta}$ cannot be shared with agent $\beta$ as well. That is to say, }

\textbf{A3:} The output of functions $Q_{\alpha}$ and $Q_{\beta}$ \emph{can} be shared with each other under the condition that they are protected by some privacy protection mechanism. 
\ignore{
the networks of their own are complex enough and unknown to each other (the input of them is unknown to each other as well based on \textbf{A2}), such that $\alpha$ and $\beta$ cannot induce each other's networks.}

Based on \textbf{A1-A3}, we aim to learn policies $\pi^*_{\alpha}$ and $\pi^*_{\beta}$ of high-quality for agents $\alpha$ and $\beta$ by preserving privacies of their own data and models. \ignore{ which is better than the one $\alpha$ learns by itself, and $\beta$ gains a high-quality policy $\pi^*_{\beta}$ by joining in the federation, instead of ``zero'' by itself ($\beta$ cannot build policies based on its own data $\{\langle s_{\beta}, a_{\beta}\rangle\}$).  }
\ignore{
In this paper, we suppose there are two agents in an environment, namely $\alpha$ and $\beta$, respectively. We would like to operate the FRL task under the following constraints:
\begin{itemize}
\item \textbf{\emph{Partial Observation}}: This is a fundamental setting in a POMDP. At each time step $t$, the full state of the environment, namely $s^t$ is unknown to each agent.  Agent $\alpha$ observes $s_\alpha^t$ and agent observes $s_\beta^t$.

\item \textbf{\emph{Different Receptivity}}: The abilities of receiving the feedback from the environment vary from agent to agent. Without loss of generality, we suppose that agent $\alpha$ cannot receive the instant reward while agent $\beta$ can. $r^t = R(s^t, a^t)$ is the instant reward at time $t$ based on the full state $s_t$ and the joint action $a_t = \{a_\alpha^t, a_\beta^t\} \in A$.

\item \textbf{\emph{Limited Communication}}: These two agents cannot directly share their own observations with each other because the data is privacy sensitive or large in size, but they need the information from each other to build a complete model. 
\end{itemize}

The goal of FRL is to learn a joint model that takes advantage of the information, i.e. the states and rewards, from both agents without sharing them, which guarantees the data privacy and the cost of communication. \hankz{Give an example of the problem here!}
}

\section{Our {\ours} Approach}
\ignore{
In value-based reinforcement learning methods, e.g. DQN \cite{Mnih2015Human}, the value function $V(s)$ (or action-value function $Q(s, a)$) can be seen as a mapping operation that encodes the states (or states and actions). A simple linear value function is not a good choice since it can be decoded by guessing inputs and weights. Therefore, we adopt neural networks which have complicated structures and some non-linear operations to encode the observations. 
We adopt the deep Q-networks \cite{Mnih2015Human} which is an extension of Q-Learning in reinforcement learning. }
In this section we present our {\ours} framework in detail. An overview of {\ours} is shown in Figure \ref{alpha-beta}, where the left part is the model of agent $\alpha$, and the right part is the model of agent $\beta$.  Each model is composed of a local Q-network with parameters $\theta_\alpha$ for agent $\alpha$, $\theta_\beta$ for agent $\beta$, and a global MLP module with parameters $\theta_g$ for both $\alpha$ and $\beta$. 

\begin{figure}
\centering
\includegraphics[width=0.49\textwidth]{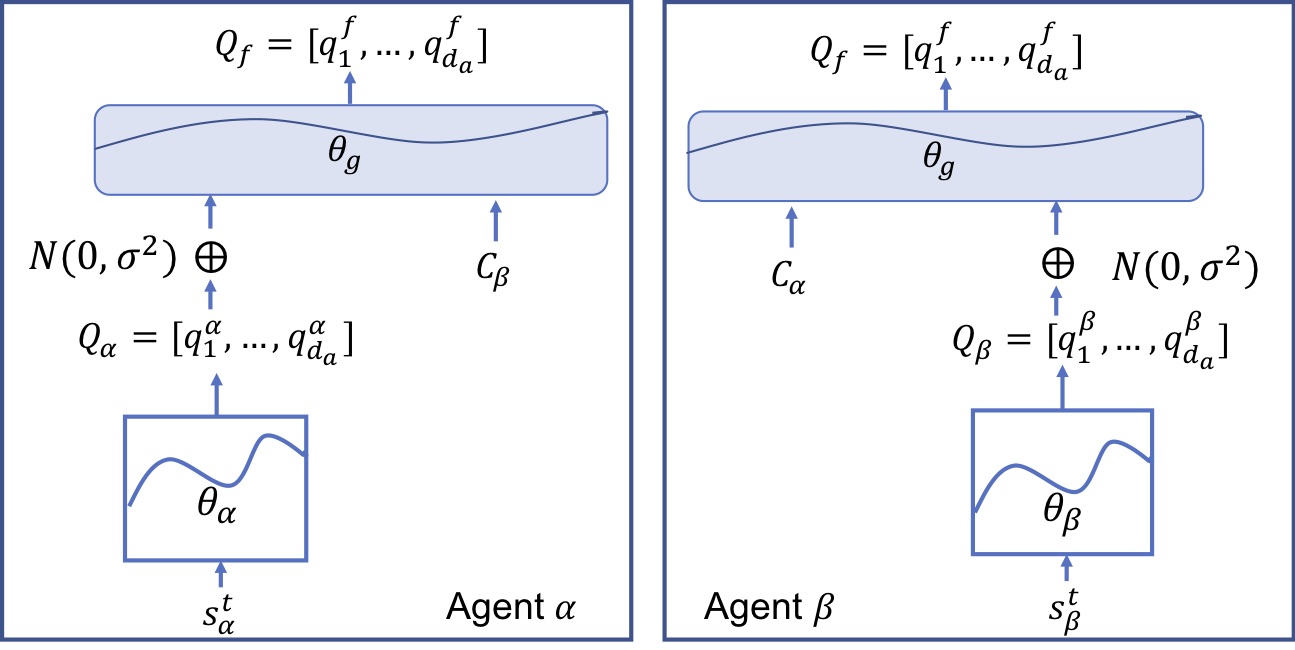}
         \caption{The networks of agents $\alpha$ and $\beta$}
         \label{alpha-beta}
\end{figure}

\ignore{
\begin{figure}
     \centering
     \begin{subfigure}[b]{0.46\textwidth}
         \centering
         \includegraphics[width=\textwidth]{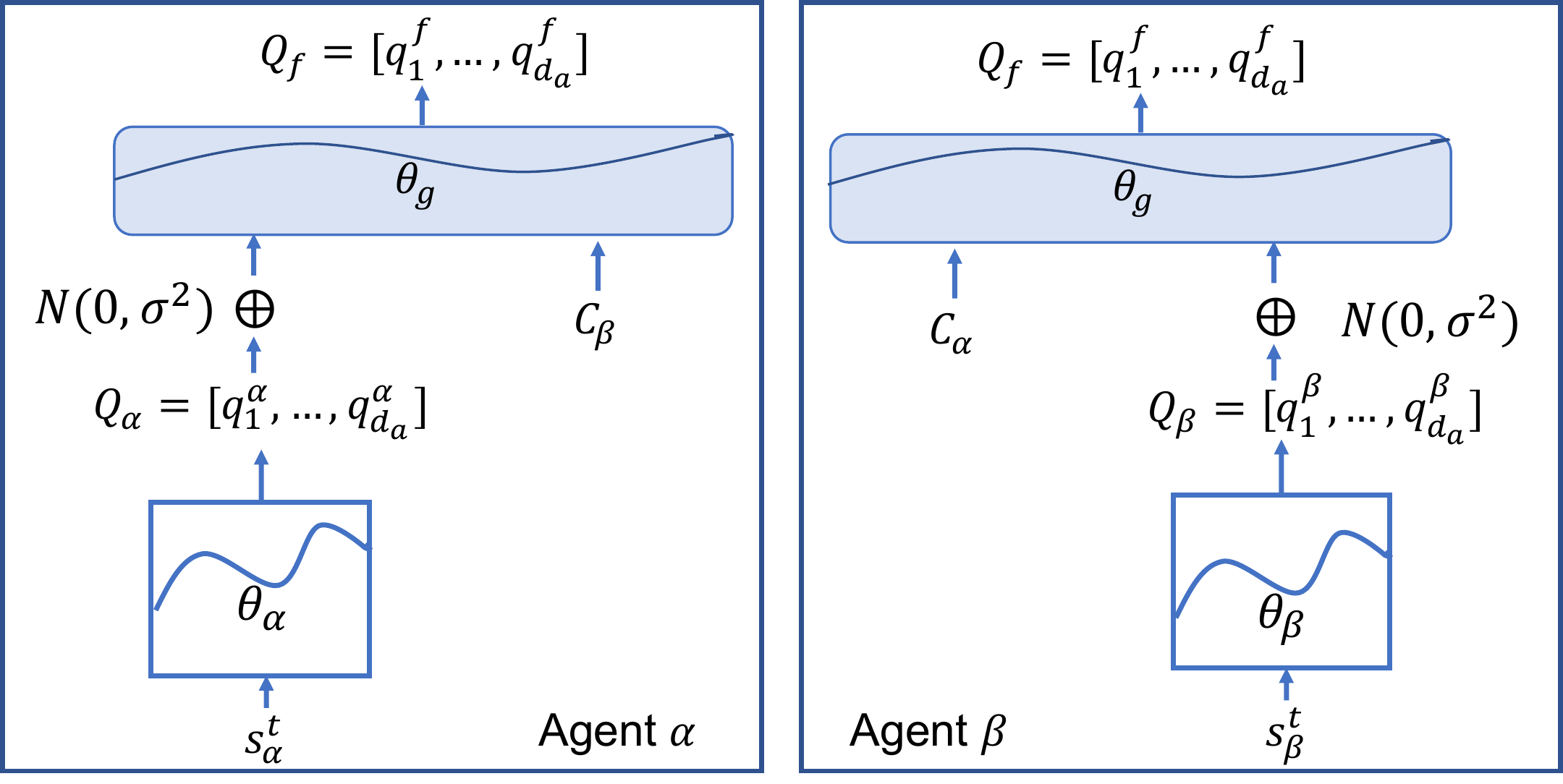}
         \caption{The networks of agents $\alpha$ and $\beta$}
         \label{alpha-beta}
     \end{subfigure}
     \hfill
     \begin{subfigure}[b]{0.53\textwidth}
         \centering
         \includegraphics[width=\textwidth]{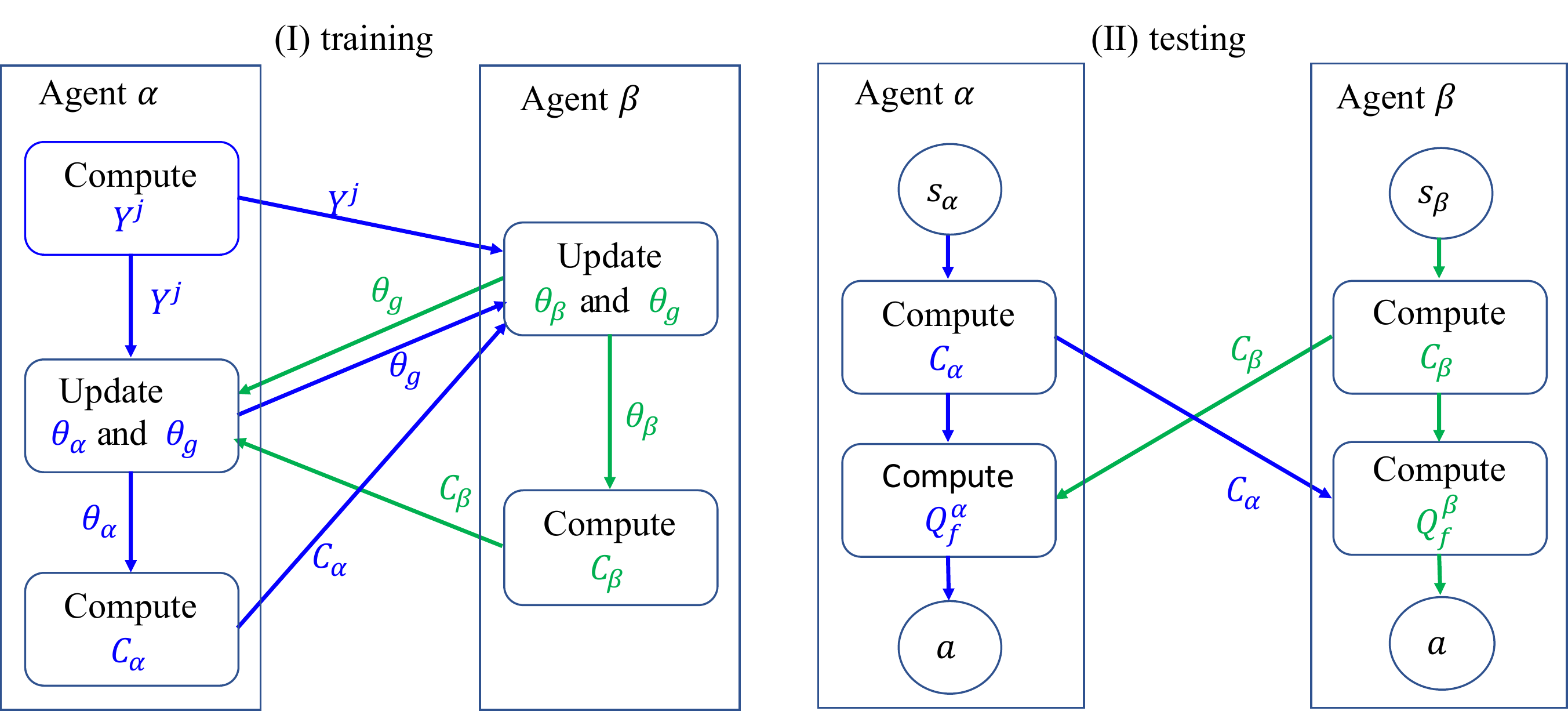}
         \caption{The training and testing procedures of agents $\alpha$ and $\beta$}
         \label{train-test}
     \end{subfigure}
\end{figure}
}
\ignore{
\begin{figure}[!ht]
\centering
\includegraphics[width=0.39\textwidth]{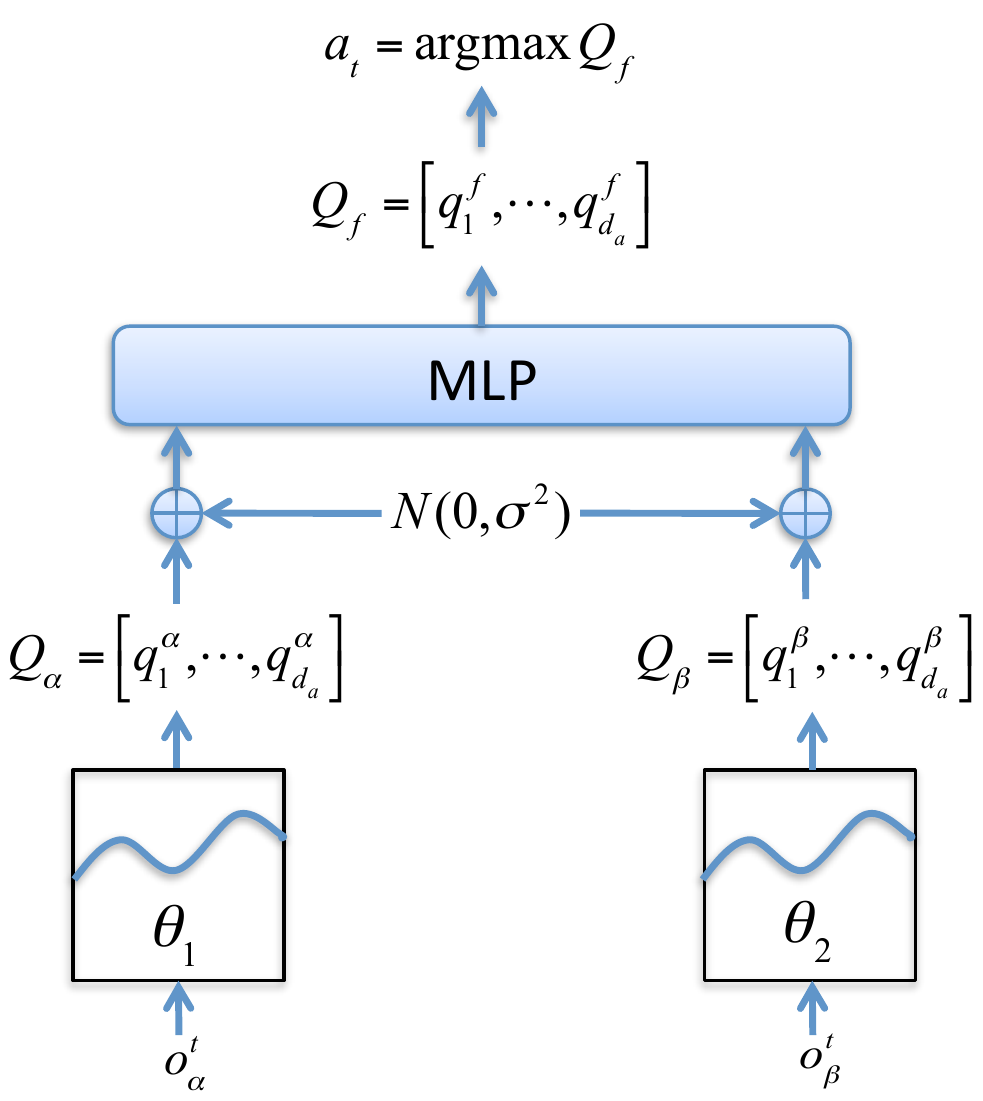}
\caption{{\ours} framework. $s_\beta^t, s_\beta^t$ are observations of agents at time step $t$. $\theta_\alpha, \theta_\beta$ are parameters of Q-networks. $Q_\alpha, Q_\beta$ are two real-valued vectors which indicate the basic q-values corresponding to the two Q-networks, $d_a$ is the dimension of the action space. $N(0, \sigma^2)$ is the Gaussian noise. $Q_f$ is the federated Q-values and $a_t$ is the predicted action at $t$.}
\label{framework}
\end{figure}
}

\paragraph{Basic Q-networks}
We build two Q-networks for agents $\alpha$ and $\beta$, denoted by $Q_\alpha(s_\alpha, a_\alpha; \theta_\alpha)$ and $Q_\beta(s_\beta, a_\beta; \theta_\beta)$, respectively, where $\theta_\alpha$ and $\theta_g$ are parameters of the Q-networks. The outputs of these two basic Q-networks are not directly used to predict the actions, but taken as input of the MLP module. 

\paragraph{Gaussian differential privacy} To avoid agents ``inducing'' models of each other according to repeatedly received outputs of other's Q-network during training, we consider using differential privacy \cite{DBLP:journals/fttcs/DworkR14} to protect the output of agent's Q-network. There are various mechanisms of differential privacy such as the Gaussian mechanism \cite{DBLP:conf/ccs/AbadiCGMMT016} and Binomial mechanism \cite{DBLP:conf/nips/AgarwalSYKM18}. In this paper, we exploit Gaussian mechanism since the output of the MLP with Gaussian input is Gaussian itself. In previous federated learning settings, the mechanism is applied to the gradients of agents (clients) before being sent to the server or other agents (clients). In our {\ours} framework, we send the output of one agent's Q-network to another, so the Gaussian noise is added to the output rather than the gradients. The mechanism is defined by
\begin{eqnarray}
\hat{Q}_\alpha(s_\alpha, a_\alpha; \theta_\alpha) = Q_\alpha(s_\alpha, a_\alpha; \theta_\alpha) + N(0, \sigma^2) \label{add noise} \\
\hat{Q}_\beta(s_\beta, a_\beta; \theta_\beta) = Q_\beta(s_\beta, a_\beta; \theta_\beta) + N(0, \sigma^2)  
\end{eqnarray}
where $N(0, \sigma^2)$ is the Gaussian distribution with mean 0 and standard deviation $\sigma$.

\paragraph{Federated Q-network} We build a new Q-network, namely \emph{federated} Q-network denoted by $Q_f$, to leverage the outputs of the two basic Q-networks with Gaussian noise based on MLP, which is defined by
\begin{eqnarray}
 && Q_f(\cdot; \theta_\alpha, \theta_\beta, \theta_g)  =   \notag \\ 
 && MLP([\hat{Q}_\alpha(s_\alpha, a_\alpha; \theta_\alpha)|\hat{Q}_\beta(s_\beta, a_\beta; \theta_\beta)];\theta_g)  \label{federated}
\end{eqnarray}
where $\theta_g$ is the parameters of MLP and $[\cdot | \cdot]$ indicates the concatenation operation. Note that parameters of an MLP can be shared between agents. Once MLP is updated, the updated parameters are shared with the other agent.

\ignore{
\fengwf{To protect the data privacy, a common way is to consider the \textbf{differential privacy} mechanism \cite{DBLP:journals/fttcs/DworkR14}, i.e. add noise to the values (e.g. gradients) before transmission. A popular mechanism is the Gaussian mechanism \cite{DBLP:conf/ccs/AbadiCGMMT016} since the sum of Gaussian is Gaussian itself. There is also another way which applies the Binomial mechanism \cite{DBLP:conf/nips/AgarwalSYKM18}.}
}

\ignore{
Our {\ours} approach can be seen as a linear combination of $Q_\alpha(s_\alpha, a; \theta_\alpha)$ and $Q_\beta(s_\beta, a; \theta_\beta)$,
\begin{equation}
Q_f(s_\alpha, s_\beta, a; \theta_\alpha, \theta_\beta) = \lambda Q_\alpha(s_\alpha, a; \theta_\alpha) + (1 - \lambda)Q_\beta(s_\beta, a; \theta_\beta),
\label{federated}
\end{equation}
where $\lambda$ is a hyper-parameter that controls the importance of Q-values of both agents. Figure \ref{framework} is an illustration of our approach. 
}

\ignore{
The federated Q-function $Q_f$ does not directly take the observations $s_\alpha$ and $s_\beta$ as inputs, but the Q-values of both Q-networks. The Q-values $Q_\alpha, Q_\beta \in R^{d_a}$, where $d_a$ is the dimension of the joint action space $A$. Although the two agents share the output Q-values with each other, they cannot  decrypt the original information of observations $s_\alpha, s_\beta$, because the structures of Q-networks are unshared and can be different from each other. The weights and input states of the Q-networks cannot be deduced by output Q-values. 
}
\ignore{
We assume that agents $\alpha$ and $\beta$ share their Q-values using Gaussian differential privacy and MLP modules but have their own private basic Q-networks. Due to Gaussian differential privacy protection and private basic Q-networks (both weights and structures of Q-networks are unknown to each other), Agents $\alpha$ and $\beta$ are unable to decode the Q-networks of each other when learning parameters $\theta_\alpha$, $\theta_\beta$ and $\theta_g$. Specifically, Agent $\alpha$ updates its own parameters $\theta_\alpha$ and parameters $\theta_g$ of MLP module by viewing the output of basic Q-networks of agent $\beta$ as a constant, while agent $\beta$ updates its own parameters $\theta_\beta$ and parameters $\theta_g$ of MLP module by viewing the output of basic Q-networks of agent $\alpha$ as a constant. We thus define the final outputs of Q-values with respect to agents $\alpha$ and $\beta$ are:}
With respect to agents $\alpha$ and $\beta$, we define each agent's federated Q-networks by viewing the other agent's basic Q-network (with Gaussian noise) as a fixed constant when updating its own basic Q-network, as shown below,
\begin{eqnarray}
Q_f^\alpha(\cdot,C_\beta;\theta_\alpha,\theta_g)=MLP([\hat{Q}_\alpha(\cdot; \theta_\alpha)| C_\beta]; \theta_g) \label{QAlpha} \\
Q_f^\beta(\cdot,C_\alpha;\theta_\beta,\theta_g)=MLP([C_\alpha | \hat{Q}_\beta(\cdot; \theta_\beta)]; \theta_g) \label{QBeta}
\end{eqnarray}
where $C_{\alpha} = \hat{Q}_\alpha(s_\alpha, a_\alpha; \theta_\alpha)$ and $C_{\beta} = \hat{Q}_\beta(s_\beta, a_\beta; \theta_\beta)$ are fixed constants when updating agent $\beta$'s basic Q-network and agent $\alpha$'s basic Q-network, respectively. 

\ignore{
To estimate the optimal action-value function, the federated framework performs value iteration, and Q-values are updated by iteratively applying Bellman updates,
\begin{eqnarray}
Q_f^\alpha(s_\alpha, a_\alpha,C_\beta; \theta_\alpha,\theta_g) = r +  \gamma \max_{a'} Q_f^\alpha(s'_\alpha, a'_\alpha,C_\beta; \theta_\alpha,\theta_g)  \label{QAlphaUpdate} \\
Q_f^\beta(s_\beta, a_\beta,C_\alpha; \theta_\beta,\theta_g)  = r +  \gamma \max_{a'} Q_f^\beta(s'_\beta, a'_\beta,C_\alpha; \theta_\beta,\theta_g)  \label{QBetaUpdate}
\end{eqnarray}
}

The Q-networks are trained by minimizing the square error loss $L^j_\alpha(\theta_\alpha,\theta_g)$ and $L^j_\beta(\theta_\beta,\theta_g)$ of agents $\alpha$ and $\beta$,
\begin{eqnarray}
L^j_\alpha(\theta_\alpha,\theta_g) = \mathbb{E} \bigg[ (Y^j -  Q_f^\alpha(s_\alpha^j, a_\alpha^j,C_{\beta}; \theta_\alpha,\theta_g) )^2 \bigg] \label{loss} \\
L^j_\beta(\theta_\beta,\theta_g) = \mathbb{E} \bigg[ (Y^j -  Q_f^\beta(s_\beta^j, a_\beta^j,C_\alpha; \theta_\beta,\theta_g) )^2 \bigg] \label{loss2}
\end{eqnarray}
where $Y^j = r^j + \gamma \max \limits_a Q_f^\alpha(s_\alpha^j, a,C_\beta; \theta_\alpha,\theta_g)$. Note that agent $\beta$ is unable to compute $Y^j$ since it does not have reward $r$. $Y^j$ is computed by agent $\alpha$ and shared with agent $\beta$. 
 
\begin{figure}[!ht]
\centering
         \includegraphics[width=0.48\textwidth]{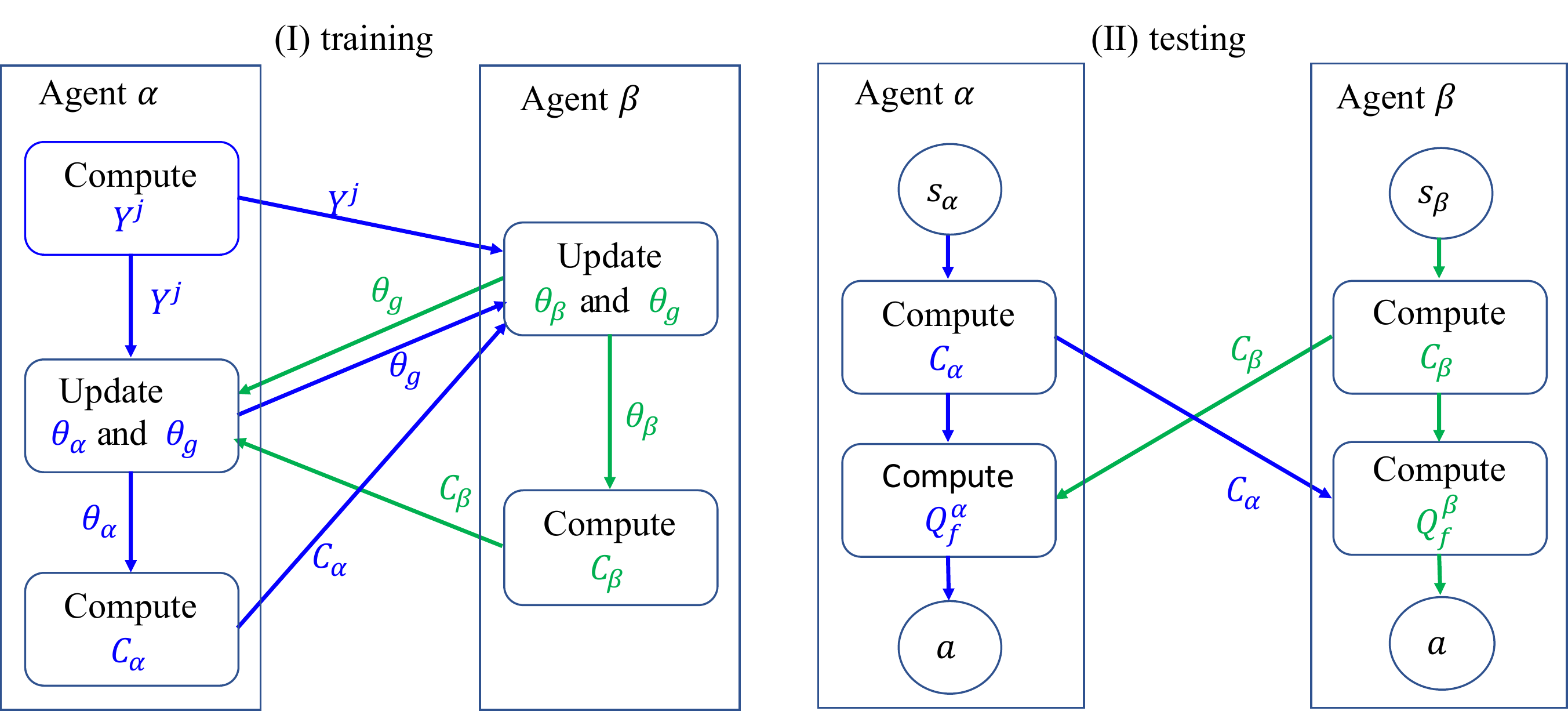}
         \caption{The training and testing procedures}
         \label{train-test}
\end{figure}
\paragraph{Overview of training and testing}
The training and testing procedures of agents $\alpha$ and $\beta$ are shown in Figure \ref{train-test}. When training, agent $\alpha$ computes $Y^j$ and sends it to agent $\beta$. Agent $\beta$ updates $\theta_\beta$ and $\theta_g$, computes $C_\beta$, and sends $\theta_g$ and $C_\beta$ to agent $\alpha$. After that, agent $\alpha$ updates $\theta_\alpha$ and $\theta_g$, computes $C_\alpha$, and sends $\theta_g$ and $C_\alpha$ to agent $\beta$. When testing, both agents compute and send $C_\alpha$ and $C_\beta$ to each other for computing $Q_f^\alpha$ and $Q_f^\beta$. 

The detailed training procedure can be seen from Algorithms \ref{FRL-alpha} and \ref{FRL-beta}. In Steps 1 and 2 of Algorithm \ref{FRL-alpha}, we initialize the basic Q-network and replay memory of agent $\alpha$ and the MLP module. In Step 3 of Algorithm \ref{FRL-alpha}, we call the function from  Algorithm \ref{FRL-beta} to initialize the Q-network and replay memory of agent $\beta$. In Step 6 of Algorithm \ref{FRL-alpha}, we obtain observation of agent $\alpha$'s state $s_\alpha^t$. In Step 7 of Algorithm \ref{FRL-alpha}, we call the function in Algorithm \ref{FRL-beta} to calculate the output of basic Q-network of agent $\beta$ and obtain  observation of agent $\beta$ and select the corresponding action, as shown in Steps from 6 to 10 of Algorithm \ref{FRL-beta}. 
In Steps from 8 and 11 of Algorithm \ref{FRL-alpha}, we perform the $\epsilon$-greedy exploration and exploitation, obtain new observations and store the transitions to the replay memory. In Steps 12 and 13 of Algorithm \ref{FRL-alpha}, we sample a record $j$ in the memory and call the function $ComputeQBeta(j)$ of Algorithm \ref{FRL-beta} to calculate the output of basic Q-network of agent $\beta$ based on the index $j$. In Steps 14 and 15 of Algorithm \ref{FRL-alpha}, we update parameters $\theta_\alpha$ and $\theta_g$. In Steps 16 and 17 of Algorithm \ref{FRL-alpha}, we compute the output of basic Q-network of agent $\alpha$ and pass it to agent $\beta$ and call the function $UpdateQ$ of agent $\beta$ to update basic Q-network of agent $\beta$ and MLP, as shown in Steps from 19 to 21 of Algorithm \ref{FRL-beta}. Note that Algorithms \ref{FRL-alpha} and \ref{FRL-beta} are executed by agents $\alpha$ and $\beta$ separately.  
\begin{algorithm}[!ht]
\caption{\tt {\ours}-ALPHA}
\label{FRL-alpha}
\textbf{Input:} state space $S_\alpha$, action space $A_\alpha$, rewards $r$  \\
\textbf{Output:} $\theta_\alpha$,$\theta_g$ 
\small
\begin{algorithmic}[1]
\STATE Initialize $Q_\alpha, Q_f$ with random values for $\theta_\alpha, \theta_g$
\STATE Initialize replay memory $D_\alpha$
\STATE Call {{\tt {\ours}-BETA}.$Init()$} 
\FOR{episode = 1: $M$}
	\REPEAT
		\STATE Observe $s_\alpha^t$
		\STATE Call {$C_{\beta}$=\tt {\ours}-BETA.$ComputeQBeta()$}
		\STATE Select action $a^t$ with probability $\epsilon$  
		\STATE Otherwise $a^t = \arg\max \limits_a Q_f^\alpha(s_\alpha^t, a, C_{\beta}; \theta_\alpha,\theta_g)$
		\STATE Execute action $a^t$, obtain reward $r^t$ and state $s^{t+1}$
		\STATE Observe $s_\alpha^{t+1}$, store $(s_\alpha^t, a^t, r^t, s_\alpha^{t+1})$ in $D_\alpha$
		\STATE Sample $(s_\alpha^j, a^j, r^j, s_\alpha^{j+1})$ from $D_\alpha$
		
		\STATE Call {$C_{\beta}$=\tt {\ours}-BETA.$ComputeQBeta(j)$}
		\STATE $Y^j = r^j + \gamma \max \limits_a Q_f^\alpha(s_\alpha^j, a,C_{\beta}; \theta_\alpha,\theta_g)$
		\STATE Update $\theta_\alpha, \theta_g$ according to Eq. (\ref{QAlpha}), (\ref{loss})
		\STATE $C_{\alpha} = \hat{Q}_\alpha(s_\alpha^j, a; \theta_\alpha)$
		\STATE Call {$\theta_g$=\tt {\ours}-BETA.$UpdateQ(Y^j,j,C_{\alpha},\theta_g)$}

	\UNTIL{terminal $t$}
\ENDFOR
\end{algorithmic}
\end{algorithm}
\begin{algorithm}[!ht]
\caption{\tt FedRL-BETA}
\label{FRL-beta}
\small
\textbf{Input:} state space $S_{\beta}$, action space $A_{\beta}$  \\
\textbf{Output:} $\theta_\beta$, $\theta_g$
\begin{algorithmic}[1]
\FUNCTION {$Init()$}{}
\STATE Initialize $Q_\beta$ with random values for $\theta_\beta$
\STATE Initialize replay memory $D_\beta$
\ENDFUNCTION
\FUNCTION{$ComputeQBeta()$}{}
\STATE Observe $s_\beta$
\STATE Select $a_\beta\in A_{\beta}$ with probability $\epsilon$  
\STATE Otherwise $a_{\beta} = \arg\max \limits_{a_{\beta}} Q_\beta(s_\beta, a_\beta; \theta_\beta)$
\STATE Store $(s_\beta,a_\beta)$ in $D_\beta$
\STATE Let $C_\beta=\hat{Q}_\beta(s_\beta, a; \theta_\beta)$
\STATE \textbf{return} $C_\beta$
\ENDFUNCTION
\FUNCTION{$ComputeQBeta(j)$}{}
\STATE Select $(s_\beta,a_\beta)$ from $D_\beta$ based on index $j$
\STATE Let $C_\beta=\hat{Q}_\beta(s_\beta, a_\beta; \theta_\beta)$
\STATE \textbf{return} $C_\beta$
\ENDFUNCTION
\FUNCTION{$UpdateQ(Y^j,j,C_{\alpha},\theta_g)$}{}
\STATE Select $(o^j_\beta,a^j_\beta)$ from $D_\beta$ based on index $j$
\STATE Update $\theta_\beta$, $\theta_g$ based on Eq. (\ref{QBeta}), (\ref{loss2})
\STATE \textbf{return} $\theta_g$
\ENDFUNCTION
\end{algorithmic}
\end{algorithm}

\ignore{
\begin{algorithm}[!ht]
\caption{\tt {\ours}-ALPHA}
\label{FRL-alpha}
\textbf{Input:} state space $S_\alpha$, action space $A_\alpha$, rewards $r$  \\
\textbf{Output:} $\theta_\alpha$,$\theta_g$ 

\begin{algorithmic}[1]
\STATE Initialize $Q_\alpha, Q_f$ with random values for $\theta_\alpha, \theta_g$
\STATE Initialize replay memory $D_\alpha$
\STATE Call {{\tt {\ours}-BETA}.$Init()$} 
\FOR{episode = 1: $M$}
	\REPEAT
		\STATE Observe $s_\alpha^t$
		\STATE Call {$C_{\beta}$=\tt FRL-BETA.$ComputeQBeta()$}
		\STATE Select action $a^t$ with probability $\epsilon$  
		\STATE Otherwise $a^t = \arg\max \limits_a Q_f^\alpha(s_\alpha^t, a, C_{\beta}; \theta_\alpha,\theta_g)$
		\STATE Execute action $a^t$, obtain reward $r^t$ and state $s^{t+1}$
		\STATE Observe $s_\alpha^{t+1}$, store $(s_\alpha^t, a^t, r^t, s_\alpha^{t+1})$ in $D_\alpha$
		\STATE Sample $(s_\alpha^j, a^j, r^j, s_\alpha^{j+1})$ from $D_\alpha$
		
		\STATE Call {$C_{\beta}$=\tt {\ours}-BETA.$ComputeQBeta(j)$}
		\STATE $Y^j = r^j + \gamma \max \limits_a Q_f^\alpha(s_\alpha^j, a,C_{\beta}; \theta_\alpha,\theta_g)$
		\STATE Update $\theta_\alpha, \theta_g$ according to Eq. (\ref{QAlpha}), (\ref{QAlphaUpdate}), (\ref{loss})
		\STATE $C_{\alpha} = \hat{Q}_\alpha(s_\alpha^j, a; \theta_\alpha)$
		\STATE Call {$\theta_g$=\tt {\ours}-BETA.$UpdateQBeta(Y^j,j,C_{\alpha},\theta_g)$}

	\UNTIL{terminal $t$}
\ENDFOR
\end{algorithmic}
\end{algorithm}

\begin{algorithm}[!ht]
\caption{\tt {\ours}-BETA}
\label{FRL-beta}
\textbf{Input:} state space $S_{\beta}$, action space $A_{\beta}$  \\
\textbf{Output:} $\theta_\beta$, $\theta_g$

\begin{algorithmic}[1]
\FUNCTION {$Init()$}{}
\STATE Initialize $Q_\beta$ with random values for $\theta_\beta$
\STATE Initialize replay memory $D_\beta$
\ENDFUNCTION
\FUNCTION{$ComputeQBeta()$}{}
\STATE Observe $s_\beta$
\STATE Select action $a_\beta\in A_{\beta}$ with probability $\epsilon$  
\STATE Otherwise $a_{\beta} = \arg\max \limits_{a_{\beta}} Q_\beta(s_\beta, a_\beta; \theta_\beta)$
\STATE store $(s_\beta,a_\beta)$ in $D_\beta$
\STATE let $C_\beta=\hat{Q}_\beta(s_\beta, a; \theta_\beta)$
\STATE \textbf{return} $C_\beta$
\ENDFUNCTION
\FUNCTION{$ComputeQBeta(j)$}{}
\STATE Select $(s_\beta,a_\beta)$ from $D_\beta$ based on index $j$
\STATE let $C_\beta=\hat{Q}_\beta(s_\beta, a_\beta; \theta_\beta)$
\STATE \textbf{return} $C_\beta$
\ENDFUNCTION
\FUNCTION{$UpdateQBeta(Y^j,j,C_{\alpha},\theta_g)$}{}
\STATE Select $(o^j_\beta,a^j_\beta)$ from $D_\beta$ based on index $j$
\STATE Update $\theta_\beta$, $\theta_g$ according to Eq. (\ref{QBeta}), (\ref{QBetaUpdate}), (\ref{loss2})
\STATE \textbf{return} $\theta_g$
\ENDFUNCTION
\end{algorithmic}
\end{algorithm}
}



\section{Experiments}
In the experiment, we evaluate {\ours}\footnote{The source code and datasets are available from https://github.com/FRL2019/FRL} in the following two aspects. Firstly, we would like to see if agent $\alpha$ can learn better policies by joining the federation than without joining the federation (agent $\alpha$ can build policies by itself since it has complete transitions $\{\langle s,a,s',r\rangle\}$, while agent $\beta$ cannot build policies without joining the federation since it has only pairs of states and actions $\{\langle s, a\rangle\}$). Secondly, we would like to see if our {\ours} approach can learn policies of high-quality which are close to the ones learnt by directly combining data of both agent $\alpha$ and $\beta$, neglecting the data-privacy issue between $\alpha$ and $\beta$. To do this, we compared our {\ours} approach to the following baselines:
\begin{itemize}
\item 
\textbf{DQN-alpha}: which is a deep Q-network \cite{Mnih2015Human} trained with agent $\alpha$'s data only. It takes observations $s_\alpha$ as input and outputs actions corresponding to $s_\alpha$. 
\item \textbf{DQN-full}: which is a deep Q-network trained by directly putting data together from both agents $\alpha$ and $\beta$, i.e., neglecting data privacy between agents $\alpha$ and $\beta$. 
\item \textbf{FCN-alpha}: which is a fully convolutional network (FCN) trained with agent $\alpha$'s data only, similar to DQN-alpha. FCN-alpha aims to build policies via supervised learning \cite{DBLP:conf/aaai/FurutaIY19}, i.e., viewing states as input and actions as labels.
\item 
\textbf{FCN-full}: which is a convolutional network trained with all data of agent $\alpha$ and $\beta$ put together directly, similar to DQN-full. 
\end{itemize}

In our experiment, we used the kernel of size $3 \times 3$ in CNN and two fully-connected layers with the size of $32\times 4$ in MLP.  We set the standard deviation in Gaussian differential privacy $\sigma$ to be 1. We adopted the Adam optimizer with learning rate 0.001 and the ReLU activation for all models. We evaluated our {\ours} with comparison to baselines in two domains, i.e., Grid-World and Text2Action. 
\subsection{Evaluation in Grid-World Domain}
\begin{figure}[!ht]
\centering
\includegraphics[width=0.26\textwidth]{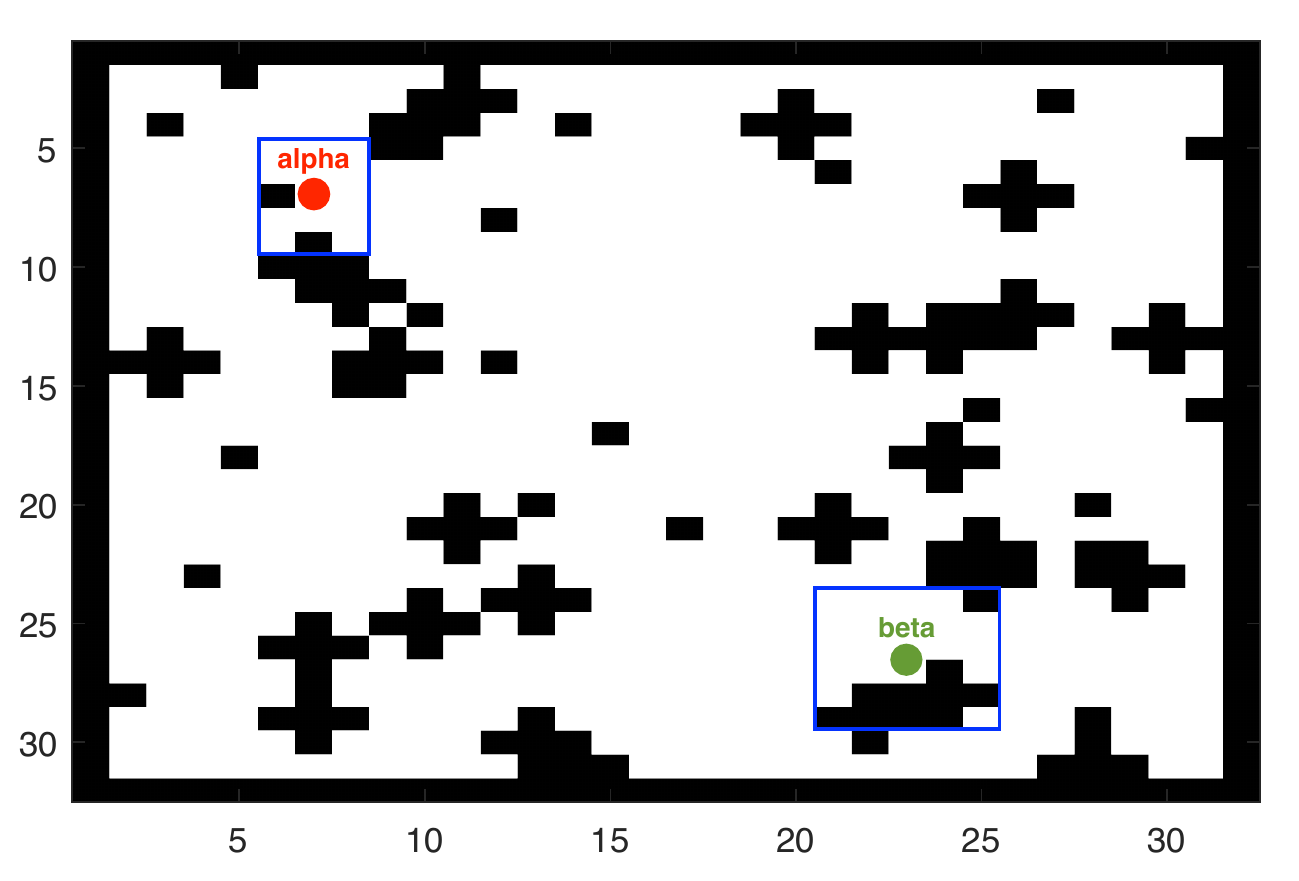}
\caption{A $32 \times 32$ grid-world domain with agents $\alpha$ and $\beta$. The rectangles in blue are boundaries of observations.}
\label{grid world example}
\end{figure}

We first conducted experiments in Grid-World domain with randomly placed obstacles \cite{DBLP:conf/nips/TamarLAWT16}. The two agents $\alpha$ and $\beta$ were randomly put in the environment as well, as shown in Figure \ref{grid world example}. The two agents aim to navigate through optimal pathes (i.e. the shortest pathes without hitting obstacles) to meet with each other. In this domain, we define states, actions and rewords as follows. 

\begin{itemize}
\item 
\textbf{States:} The domain is represented by a $N_g \times N_g$ binary-valued matrix (0 for obstacle, 1 otherwise), where $N_g$ is the size of the domain. We evaluated our approach with respect to different sizes, i.e., $N_g = 8, 16, 32$. The observed state of agent $\alpha$, denoted by $s_\alpha$, was set to be a $3 \times 3$ matrix with the current position of agent $\alpha$ in the center of the matrix. The observed state of agent $\beta$, denoted by $s_\beta$, was set to be a $5 \times 5$ matrix with the current position of agent $\beta$ in the center of the matrix. \ignore{Since the single $3 \times 3$ or $5 \times 5$ grid information may be insufficient for an agent to confirm its location and make a decision in a partially observed environment, i.e. there are many $3 \times 3$ or $5 \times 5$ grids which are the same, we adopt a sequence of observations to construct an observation, i.e. $s_\alpha^t = [s_\alpha^{t-H+1}, s_\alpha^{t-H+2}, \dots, s_\alpha^t]$, where $s_\alpha^t$ is a $3 \times 3$ grid observation at time $t$ of agent $\alpha$ and $H$ is the length of history. We calculate the average length of optimal paths for each domains, and find out that each agent need to move 2.4 steps in $8 \times 8$ grid, 4.8 steps in $16 \times 16$ grid and 9.8 steps in $32 \times 32$ grid on average. Therefore, we empirically set $H = 2, 4, 8$ for $8 \times 8, 16 \times 16, 32 \times 32$ domain, respectively.}

\item 
\textbf{Actions:} There are 4 actions for each agent, i.e. going towards 4 directions, denoted by $\{east, south, west, north\}$. 
\item 
\textbf{Rewards:} The reward is composed of two parts, i.e., local reward $r_l$ and global reward $r_g$, respectively. When an agent hits an obstacle, $r_l$ is set to be -10; when an agent meets the other agent, $r_l$ is set to be +50; otherwise, $r_l$ is set to be -1. Considering the goal of the task is to make the two agents eventually meet each other, we exploited an additional reward regarding the distance between two agents, namely global reward, i.e., $r_g = c / md(\alpha, \beta)$, where $md(\alpha, \beta)$ is the Manhattan distance between the two agents, and $c$ is a regularization factor which is set to be the dimension of the domain, i.e. $c = N_g$. The final reward $r$ is calculated by $r=r_l+r_g$.

\ignore{
\textbf{Q-Network:} We devise a CNN-based Q-Network, with a convolutional layer of 32 kernels of size $3 \times 3$ and paddings. The outputs of the convolutional layer are then flatten and fed to two fully connected layers with 256 and 4.
}
\ignore{
\textbf{Baseline:} DQN-alpha only uses the alpha Q-network. CNN-alpha has the same network structure as DQN-alpha, except that it is trained in the supervised way.  DQN-full uses both networks, i.e. $s_\alpha$ and $s_\beta$ will be input to two different convolutional layer with $3 \times 3$ sized kernels, then the outputs will be flatten, concatenated and input into the fully connected layers with size 256 and 4. CNN-full has the same network structure as DQN-full, except that it is trained in the supervised way.   \textbf{\ours} federates the two networks according to Eq (\ref{add noise}) - Eq(\ref{federated}), where the MLP module comprises two fully-connected layers with size 32 and 4. The noise probability $p$ is set to 0.5, $\sigma$ (the standard deviation of Gaussian noise) is set to 1. For all models, we adopt the Adam optimizer with learning rate 0.001 and the ReLU activation.
}
\item 
\textbf{Dataset:} We generated 8000 different maps (or matrices) for each size of $8 \times 8, 16 \times 16$ and $32 \times 32$. In each map, we randomly chose two positions for the two agents, and computed the optimal path (shortest path) which was compared to the paths predicted by our {\ours} and baselines. We randomly split the 8000 maps into 6400 for training, 800 for validation, and 800 for testing. 
\item
\textbf{Criteria:} In each training episode, we first took the initial state and predicted an action with a model. We then got a new state and took it as the new input of the model at the second time step. We repeated the procedure until the two agents met each other, which is indicated as \emph{a successful episode}, or it exceeded the maximum time step $T_m$, which is indicated as \emph{a failure episode}. We set $T_m$ to be twice the length of the longest optimal path, i.e. $T_m = 38, 86, 178$ for each size of $8 \times 8, 16 \times 16$ and $32 \times 32$, respectively. We finally computed the measures of successful rate $SuccRate$, and average reward $AvgRwd$\ignore{, and trajectory difference $TrajDiff$}, i.e., \[SuccRate = \frac{\# SuccessfulEpisodes}{\# TotalEpisodes},\] and \[AvgRwd = \frac{TotalCumulativeReward}{\# TotalEpisodes},\]\ignore{ and $TrajDiff = \frac{| LengthOfPredictedPaths - LengthOfShortestPaths |}{LengthOfShortestPaths}$} where $\# SuccessfulEpisodes$ indicates the number of successful episodes, $\# TotalEpisodes$ indicates the total number of episodes, $TotalCumulativeReward$ indicates the total reward of all episodes, \ignore{$LengthOfPredictedPaths$ indicates the total number of steps of predicted paths (a path is a sequence of actions predicted by a model), $LengthOfShortestPaths$ indicates the total number of steps of optimal paths, }respectively. 
\end{itemize}

\begin{table}[!ht]
\begin{minipage}[t]{0.49\textwidth}
\begin{small}
\centering
\caption{Comparison with baselines in Grid-World}
\begin{tabular}{|c|c|c|c|c|}
\hline
\multirow{2}{*}{\textbf{Metric}} & \multirow{2}{*}{\textbf{Method}} & \multicolumn{3}{c|}{\textbf{Domain} w.r.t. various sizes} \\ \cline{3-5}
              				&         & $8 \times 8$   & $16 \times 16$   & $32 \times 32$ \\ \hline
\multirow{3}{*}{$SuccRate$} & FCN-alpha  & 69.73\%             & 48.04\% 	& 41.73\% \\ \cline{2-5} 
					& DQN-alpha  & 88.27\%             & 76.20\% 	& 71.41\%  \\ \cline{2-5}
					& FedRL-1        & 92.52\%             & 79.83\% 	& 77.88\%   \\ \cline{2-5}
					& FedRL-2        & \textbf{95.06\%} & \textbf{84.31\%} & \textbf{82.02\%}   \\ \cline{2-5}
					& \grey{FCN-full} & \grey{72.16\%} & \grey{56.44\%} & \grey{50.15\%}  \\ \cline{2-5}
				       & \grey{DQN-full} & \grey{93.69\%} & \grey{83.40\%} & \grey{79.73\%}   \\ 			       
\hline \hline
\multirow{3}{*}{$AvgRwd$} 	& DQN-alpha        & 13.781             & -112.084           & -285.946   \\ \cline{2-5}
					& FedRL-1              & 18.152             & -94.193             & -226.583   \\ \cline{2-5}
					& FedRL-2               & \textbf{19.101} & \textbf{-84.139} & \textbf{-189.756}   \\ \cline{2-5}
					& \grey{DQN-full} & \grey{31.286}  & \grey{-38.114}    & \grey{-52.72}   \\
\hline 
\end{tabular}
\label{grid-world}
\end{small}
\end{minipage}
\end{table}
\paragraph{Experimental Results w.r.t. Domain Sizes} 
We ran our {\ours} and baselines five times and calculated an average of $SuccRate$ (as well as $AvgRwd$). We calculated two results of our {\ours}, denoted by {\ours}-1 and {\ours}-2, which correspond to ``adding Gausian differential privacy on the local Q-network'' and ''not adding Gausian differential privacy on the local Q-network'' in the testing data, respectively. The results are shown in Table \ref{grid-world}. From Table \ref{grid-world}, we can see that $SuccRate$ of both {\ours}-1 and {\ours}-2 is much better than DQN-alpha and CNN-alpha in all three different sizes of domains, which indicates agent $\alpha$ can indeed get help from agent $\beta$ via learning federatively with agent $\beta$. Comparing to DQN-full, we can see that the $SuccRate$ of both {\ours}-1 and {\ours}-2 is close to DQN-full, which indicates our federated learning framework can indeed take advantage of both training data from agents $\alpha$ and $\beta$, even though they are protected locally with Gausian differential privacy (the reason why {\ours}-2 is slightly better than DQN-full is that the hierarchical structure of our federated learning framework with three components may be more suitable for this domain than a unique DQN framework). In addition, we can also see that the $SuccRate$  generally decreases when the size of the domain increases, which is consistent with our intuition, since the larger the domain is, the more difficult the task is (which requires more training data to build models of high-quality). 

\ignore{DQN-based models perform better than CNN-based classification models. When considering the \emph{trajectory difference} metric, our {\ours} models outperform both FCN-alpha and FCN-full. Compared to DQN-alpha, {\ours} models induce a bit increment of trajectory difference in all domains. Meanwhile, DQN-full also induces the trajectory difference in $16 \times 16$ and $32 \times 32$ domains.
Because these models try more steps to success and get more rewards in the long run, which can be demonstrated by the results of \emph{average cumulative reward} (CNN-alpha and CNN-full do not involve in this metric). } 
From the metric of $AvgRwd$, we can see that both {\ours}-1 and {\ours}-2 are better than DQN-alpha, which indicates agent $\alpha$ can indeed get help from agent $\beta$ in gaining rewards via federated learning with agent $\beta$. We can also find that {\ours}-2 outperforms {\ours}-1 in both $AvgRwd$ and $SuccRate$. This is because our model is trained based on Gausian differential privacy, indicating $SuccRate$ on testing data with Gausian differential privacy, as done by {\ours}-2, should be better than on testing data without Gausian differential privacy, as done by {\ours}-1. 

\ignore{
Regarding the different settings of Gaussian mechanism, the FRL-1 model performs more successful episodes and gets more reward than the FRL-2 model, while it induces a little bit more trajectory difference. The result shows that our {\ours} model is more suitable to Q-values with noise than original Q-values, which means that our approach can protect data privacy better.}

\begin{figure*}[!ht]
\begin{center}
\includegraphics[width=0.28\textwidth]{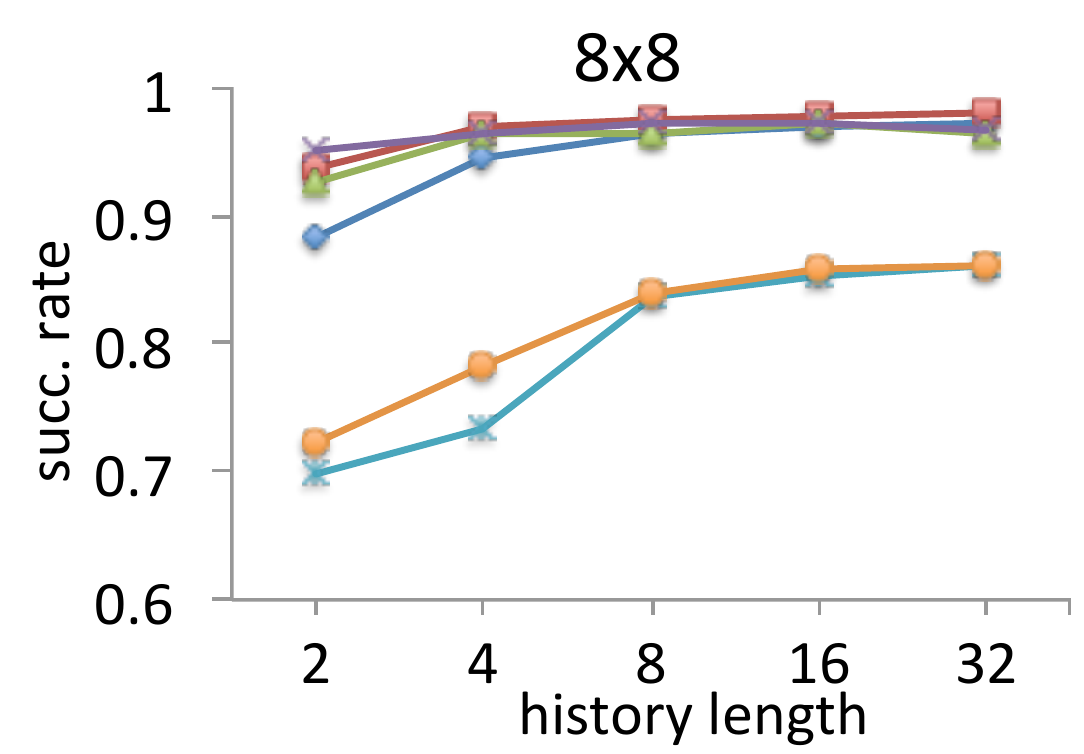}
\includegraphics[width=0.28\textwidth]{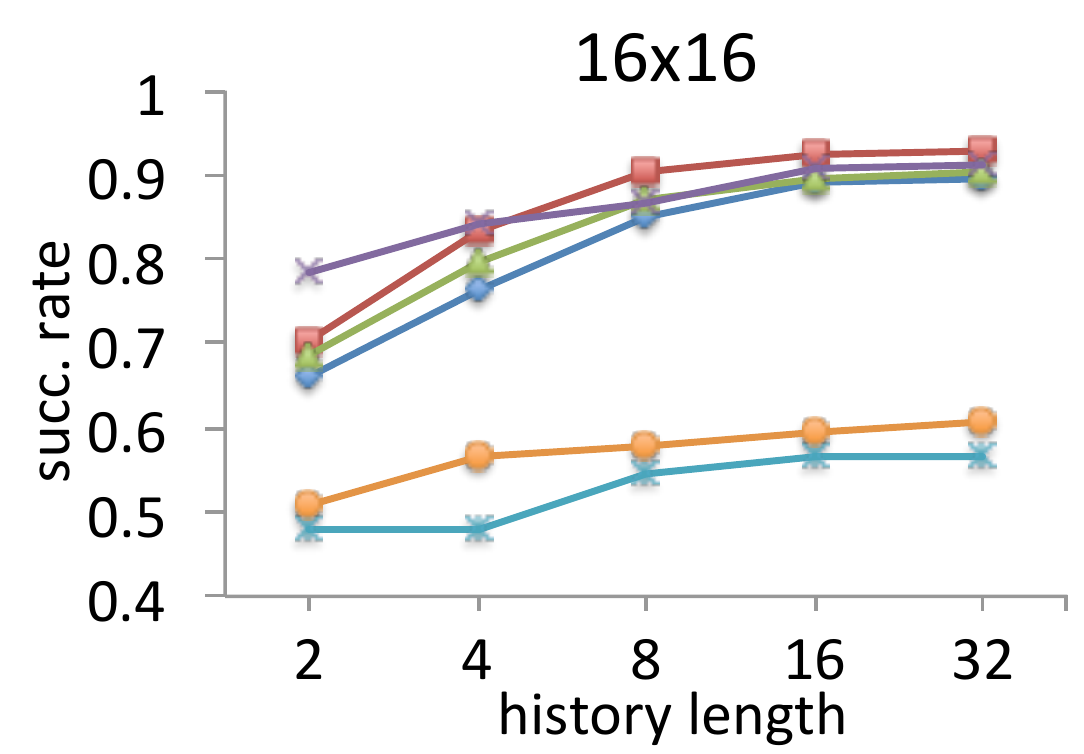}
\includegraphics[width=0.34\textwidth]{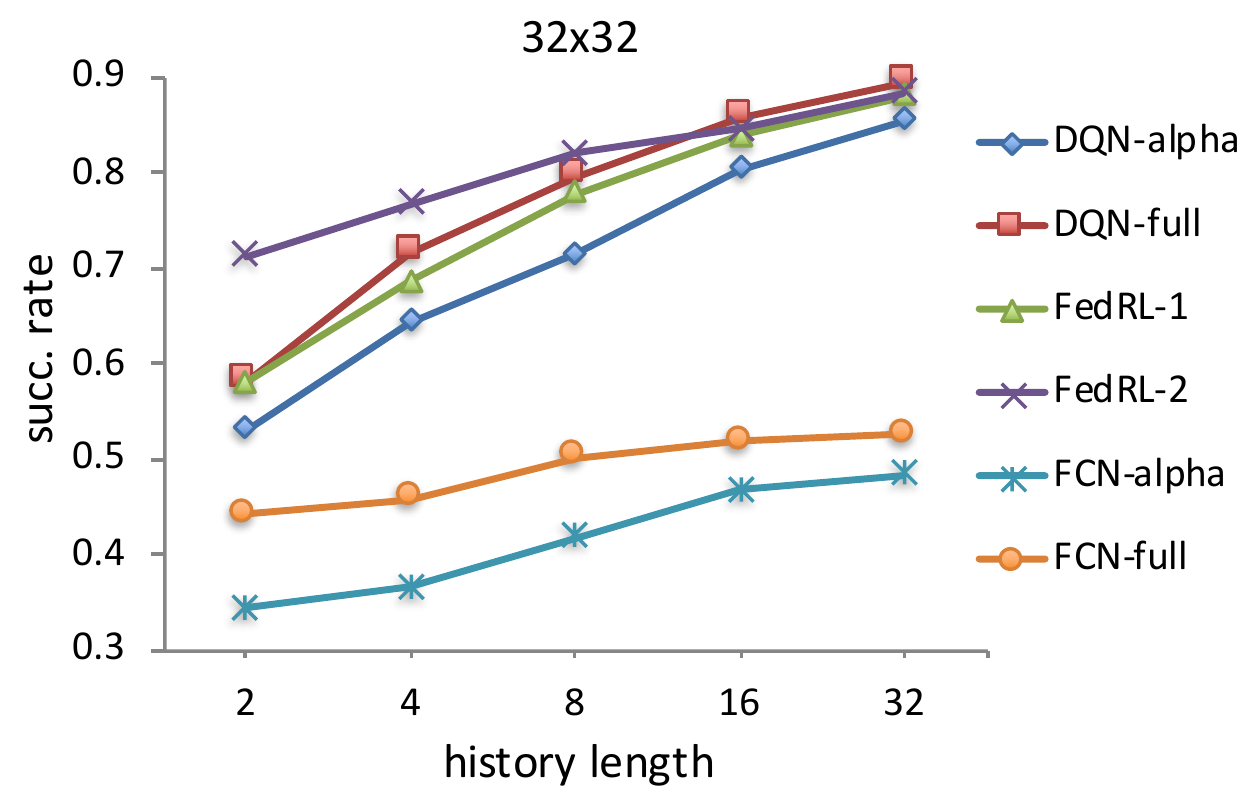}
\includegraphics[width=0.28\textwidth]{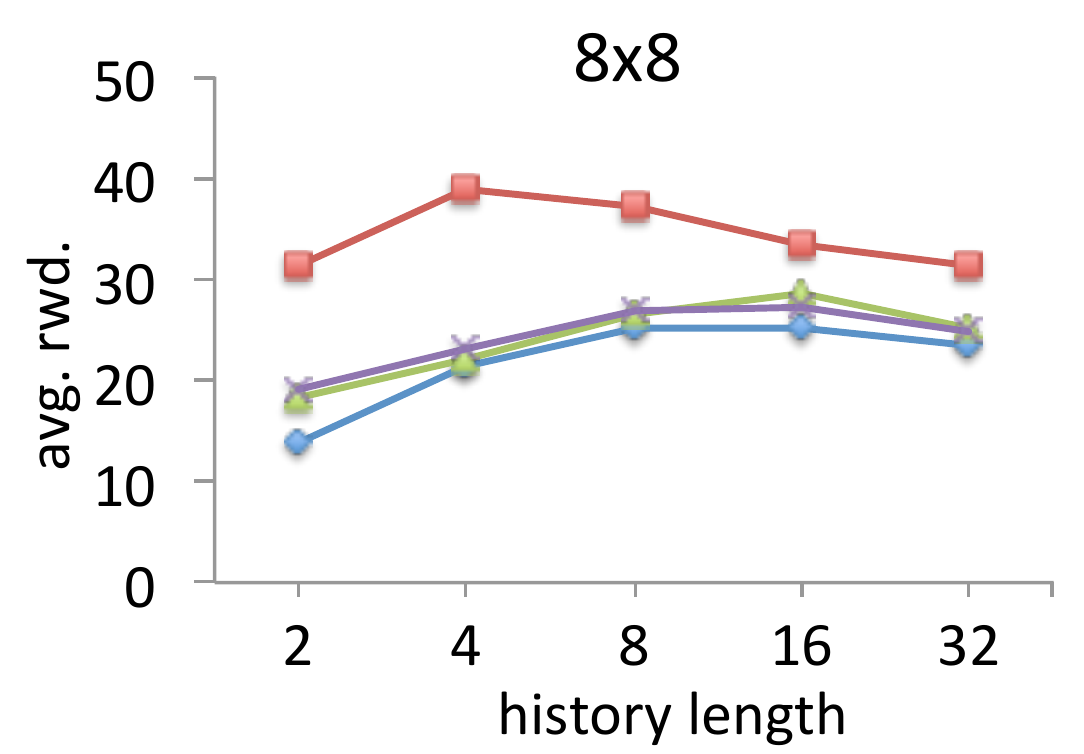}
\includegraphics[width=0.28\textwidth]{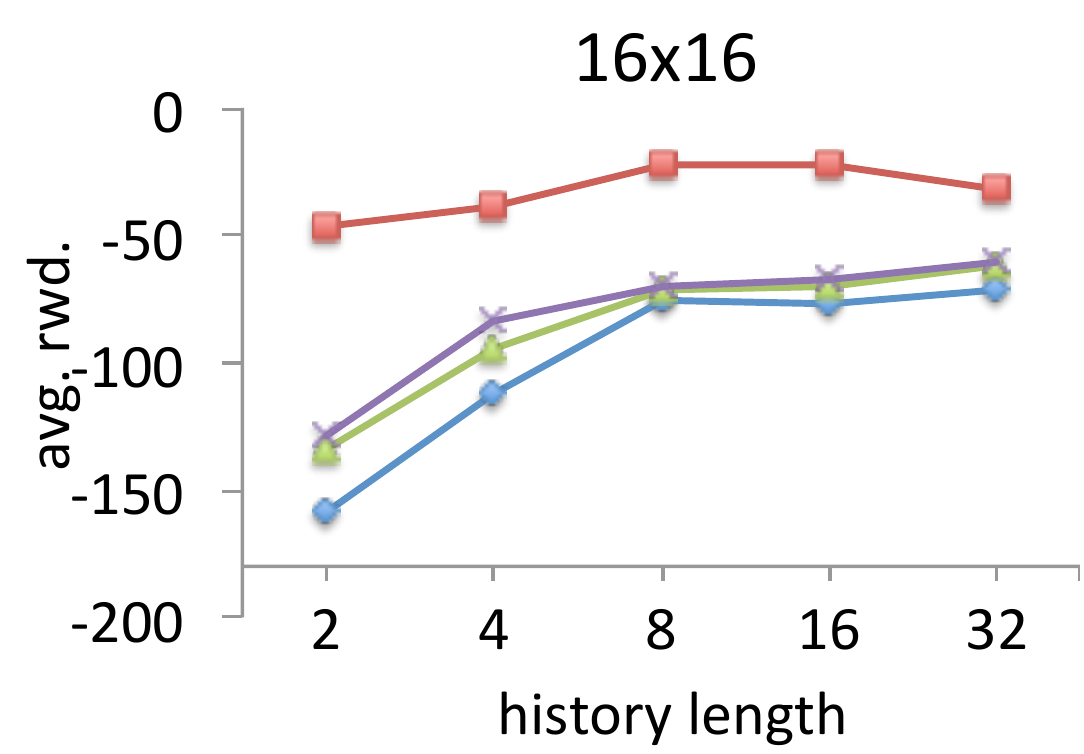}
\includegraphics[width=0.34\textwidth]{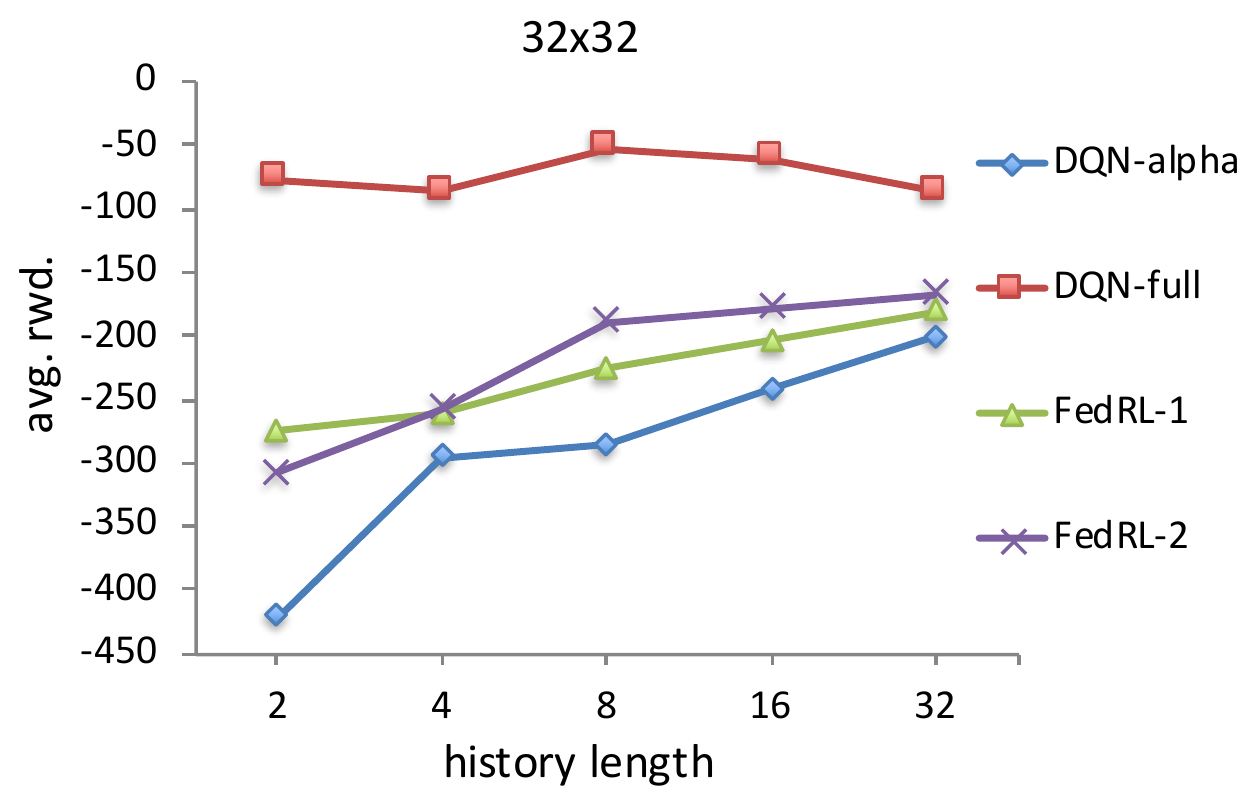}
\caption{Results about the impact of history length.}
\label{results of history length}
\end{center}
\end{figure*}
\paragraph{Experimental Results w.r.t. History Length}
To study when {\ours} works, we consider the amount of information that an agent uses. The observation input at each time is a sequence of observations, i.e. observation history. Intuitively, the longer the length of the observation history, the more information we have, and the more complicated the neural network is. We fixed the structures of all models and only changed the length of the history observations. We tested the length of history from 2 to 32 for all domains. The results are shown in Figure \ref{results of history length}. In the first row of Figure \ref{results of history length}, we can observe is that the success rates are improving with the increment of the history length. In $8 \times 8$ and $16 \times 16$ domains, the results converge when history length is longer than 16, while in $32 \times 32$ domain, they have not converged even at the history length of 32. The reason is that $32 \times 32$ domain is more complicated than the other two domains, so it needs more information (i.e. $H > 32$) to learn a model of high-quality. We can also find that when the history length is short (i.e. $H = 2$ for $16 \times 16$ and $H \leq 4$ for $32 \times 32$), DQN-alpha, DQN-full and {\ours}-1 perform poorly. {\ours}-1 and DQN-full do not show their advantages although they take as input both $s_\alpha$ and $s_\beta$ directly or indirectly. However, {\ours}-2, which applies differential privacy to both training and testing samples, shows its great scalability even with limited amount of history. In small domains, such as $8 \times 8$, a few steps are enough to explore the whole environment. Therefore, the DQN-alpha which takes only single observation as input can performs as well as DQN-alpha and {\ours} models.

\ignore{We conjecture it is because the models have not been well-learned with limited input information. The experiments demonstrate that our {\ours} can perform as well as DQN-full in all metrics.}

\subsection{Text2Actions Domain}
In this experiment, we evaluated our {\ours} in another domain, i.e., Text2Action \cite{DBLP:conf/ijcai/FengZK18}, which aims to extract action sequences from texts. For example, consider a text ``\emph{Cook the rice the day before, or use leftover rice in the refrigerator. The important thing to remember is not to heat up the rice, but keep it cold.}", which addresses the procedure of making egg fired rice. The task is to extract words composing an action sequence ``cook(rice), keep(rice, cold)'', or ``use(leftover rice), keep(rice, cold)''. \ignore{An overall process of extracting action sequences is shown in Figure \ref{eas example}.}\ignore{We assume that there are two agents, $\alpha$ and $\beta$, where agent $\alpha$ has a rich word embedding model trained by a sufficiently large corpus, agent $\beta$ has a POS-tagger that can generate the part of speech of each word from a text.} We define states, actions and rewords as follows. 

\ignore{
\begin{figure}[!ht]
\begin{center}
\includegraphics[width=0.5\textwidth]{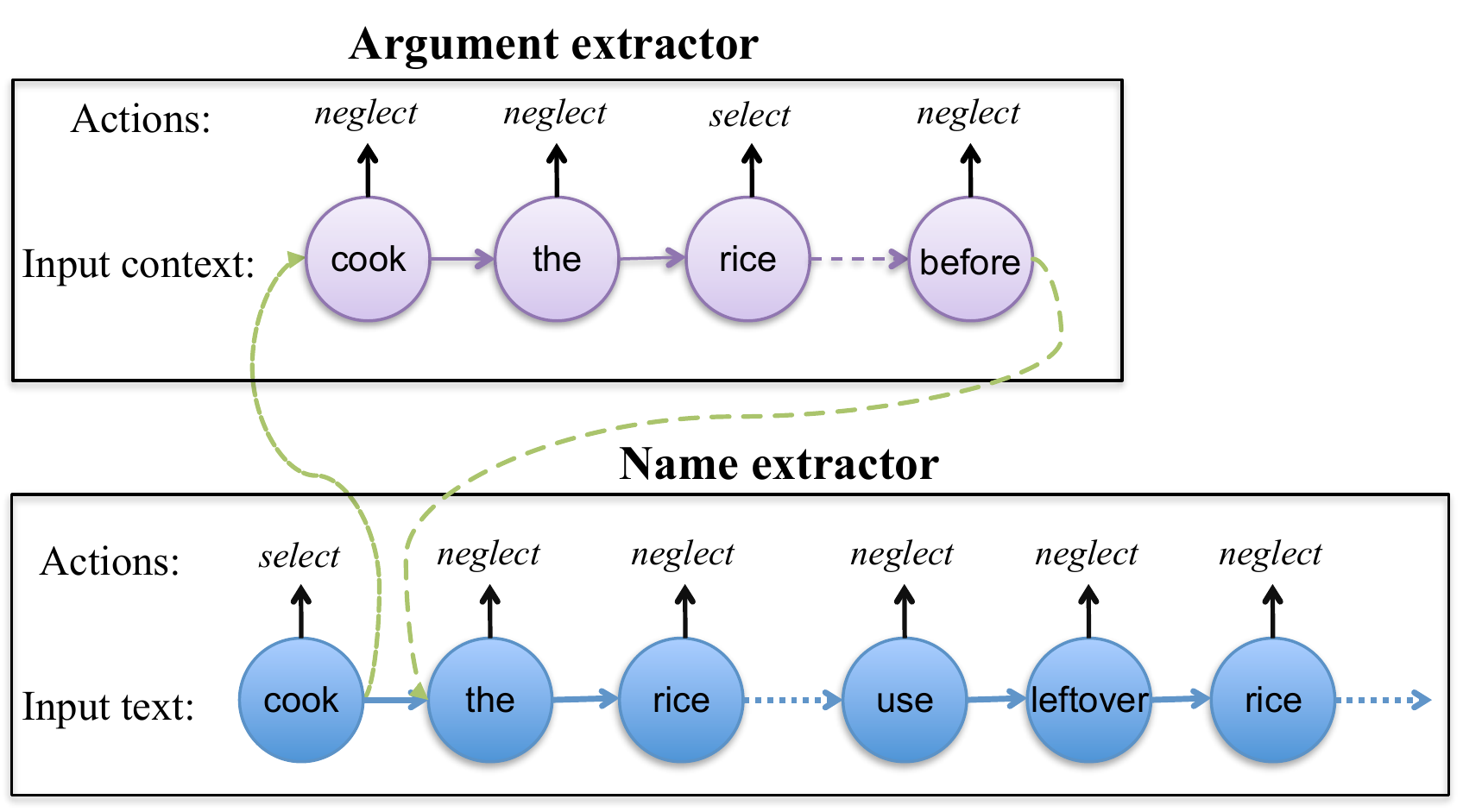}
\end{center}
\caption{Illustration of the overall process of the Text2Action task. ``Name extractor'' is a module for extracting action names. ``Argument extractor'' is a module for extracting action arguments based on the extracted action names and their context in the text.}
\label{eas example}
\end{figure}
}

\begin{table}[!ht]
\begin{small}
\caption{Comparison with baselines in Text2Action}
\begin{center}
\begin{tabular}{|c|c|c|c|c|}
\hline
\multirow{2}{*}{\textbf{Metric}} & \multirow{2}{*}{\textbf{Method}} & \multicolumn{3}{c|}{\textbf{Dataset}} \\ \cline{3-5}
              				&         		& WHS   		    & CT                      & WHG \\ \hline
\multirow{3}{*}{F1} 	& FCN-alpha  	& 86.19\%             & 65.44\%             &  55.18\%  \\ \cline{2-5}
					& DQN-alpha  	& 92.11\%             & 74.64\%              &  66.37\%  \\ \cline{2-5}
					& FedRL-1            & 93.76\%            & \textbf{85.05\%} & 75.64\%   \\ \cline{2-5}
					& FedRL-2            & \textbf{94.41\%} & 84.92\%            & \textbf{75.85\%}   \\ \cline{2-5}
				       & \grey{FCN-full} & \grey{91.42\%} & \grey{79.03\%} & \grey{68.93\%}   \\ \cline{2-5}
				       & \grey{DQN-full} & \grey{94.55\%} & \grey{83.39\%} & \grey{74.63\%}   \\ 
				       
\hline
\multirow{3}{*}{$AvgRwd$} 	& DQN-alpha       & 54.762             & 47.375            & 46.510   \\ \cline{2-5}
					& FedRL-1              & 55.623            & \textbf{50.472} & 48.359   \\ \cline{2-5}
					& FedRL-2              & \textbf{55.894} & 50.452            & \textbf{48.373}   \\ \cline{2-5}
					& \grey{DQN-full} & \grey{56.192} & \grey{50.307}  & \grey{48.154}   \\
\hline
\end{tabular}
\label{Text2Action}
\end{center}
\end{small}
\end{table}

\begin{figure*}[!ht]
\centering
\includegraphics[width=0.8\textwidth]{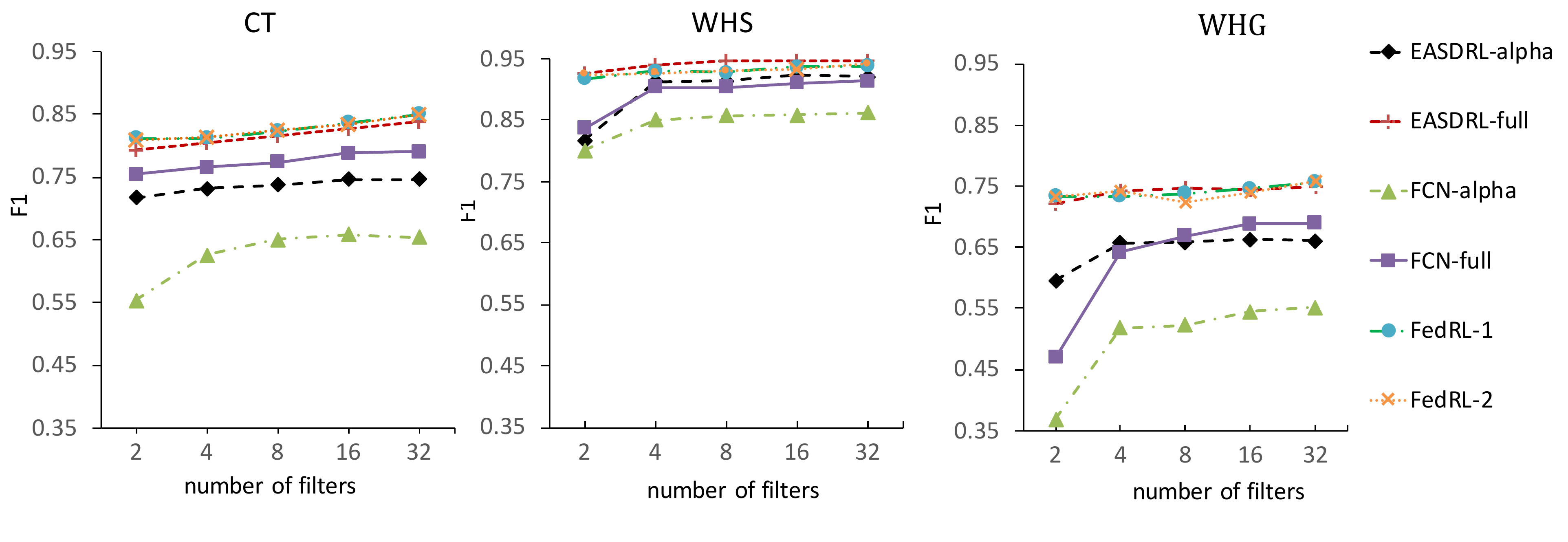}
\caption{Results about the impact of the number of convolutional kernels.}
\label{results of model complexity}
\end{figure*}

\begin{itemize}
\item
\textbf{States:} $s_\alpha \in \mathbb{R}^{N_w \times K_1}$ is real-valued matrix that describes the part-of-speech of words, and $s_\beta \in \mathbb{R}^{N_w \times K_2}$ is real-valued matrix that describes the embedding of words. $N_w$ is the number of words in the text, $K_1$ is the dimension of a part-of-speech vector, and $K_2$ is the dimension of word vectors. In our experiments, part-of-speech vectors are randomly initialized and trained together with the Q-network, while word vectors are generated from the pre-trained word embeddings and will not be changed during the training of the Q-network. 

\item 
\textbf{Actions:} There are two actions for each agent, i.e., $\{select, neglect\}$, indicating selecting a word as an action name (or a parameter), or neglecting a word (which means the corresponding word is neither an action name nor a parameter).

\item 
\textbf{Rewards:} The instant reward includes a basic reward and an additional reward, where the basic reward indicates whether the agent selects a word correctly or not, and the additional reward encodes the priori knowledge of the domain, i.e., the proportion of words that are related to action sequences \cite{DBLP:conf/ijcai/FengZK18}. We assume that only agent $\alpha$ knows the rewards, while agent $\beta$ does not know them.

\item
\textbf{Dataset:} We conducted experiments on three datasets, i.e., ``Microsoft Windows
Help and Support'' (WHS) documents \cite{DBLP:conf/acl/BranavanCZB09},
and two datasets collected from ``WikiHow Home and
Garden''\footnote{https://www.wikihow.com/Category:Home-and-Garden}
(WHG) and ``CookingTutorial''\footnote{http://cookingtutorials.com/}
(CT), respectively.  In CT, there are 116 labeled texts and 134,000 words, with 10.37\% being action names and 7.44\% being action arguments. In WHG, there are 150 labeled texts and 34,000,000 words, with 7.61\% being action names and 6.3\% being action arguments. In WHS, there are 154 labeled texts and 1,500 words, with 19.47\% being action names and 15.45\% being action arguments.

\item
\textbf{Criteria:}  For evaluation, we first fed texts to each model to obtain selected words. We then compared the outputs to their corresponding ground truth and calculated $\# TotalTruth$ (total ground truth words), $\# TotalRight$ (total correctly selected words), and $TotalSelected$ (total selected words). After that we computed metrics 
$precision = \frac{\#TotalRight}{\#TotalSelected}$, 
$recall = \frac{\#TotalRight}{\#TotalTruth}$,
and 
\[F1 = \frac{2 \times precision \times recall}{precision + recall}.\] 
We used F1-metric for all baselines. For reinforcement learning methods, we also computed the average cumulative rewards 
\[AvgRwd = \frac{TotalCumulativeReward}{\# TotalTimeSteps},\] where $TotalCumulativeReward$ indicates the total cumulative reward of all testing texts and $\# TotalTimeSteps$ indicates the total number of steps of all testing texts. 
\end{itemize}
We adopted the same TextCNN structure as \cite{DBLP:conf/ijcai/FengZK18}. Four convolutional layers corresponding to bigram, tri-gram, four-gram and five-gram with 32 kernels of size $n \times m$, where $n$ refers to the $n$-gram and $m = 50$. Each convolutional layer is followed by a max-pooling layer with size of $(N_w - n + 1, 1)$, where $N_w - n + 1$ is the first dimension of the outputs of the n-gram convolutional layer. The max-pooling outputs are concatenated and fed to two fully connected layers with size $128\times 2$, where $128$ equals to $4 \times 32$, indicating 4 types of $n$-grams and 32 kernels for each $n$-grams, and 2 is the size of the action space. The MLP module of {\ours} is composed of two fully-connected layers with size $4\times 2$. The standard deviation of Gaussian differential privacy $\sigma$ was set to be 1. For all models, we adopted the Adam optimizer with learning rate 0.001 and ReLU activation\footnote{Detailed setting can be found from the source code: https://github.com/FRL2019/FRL}.

\paragraph{Experimental Results}
We ran our {\ours} and baselines five times to calculate an average of F1 (as well as $AvgRwd$). The results are shown in Table \ref{Text2Action}. From Table \ref{Text2Action} we can see that both {\ours}-1 and {\ours}-2 outperform both FCN-alpha and DQN-alpha in all three datasets under both F1 and $AverRwd$ metrics, which indicates agent $\alpha$ can learn better policies via federated learning with agent $\beta$ than learning by itself. Comparing our {\ours}-1 and {\ours}-2 with DQN-full in both F1 and $AvgRwd$, we can see that their performances are close to each other, suggesting our federated learning framework performs as good as the DQN model directly built from all training data from both agents $\alpha$ and $\beta$.

\ignore{
In addition, 
Texts of the WHS dataset are quite short and most verbs of the texts are exactly the words of actions to be selected, which makes the Text2Action task easy to solve, i.e. WHS is a naive domain. Therefore, all baselines perform well in WHS, and our {\ours} only outperforms the DQN-alpha, CNN-full and CNN-alpha by around $2\% \sim 8\%$. When processing more complicated texts which contain much more redundant verbs and sentences, e.g. texts in CT and WHG datasets, the performance of our {\ours} model is dominant, improving the F1 score around $10\% \sim 20\%$ absolutely compared to the baselines. Under all metrics, our {\ours} performs as well as DQN-full. Generally speaking, {\ours}-2 excels {\ours}-1, but the advantage is not so obvious.
}


To see the impact of the model complexity, we varied the number of convolutional kernels from 2 to 32 \ignore{(the last fully connected layer will change from $8 = 4 \times 2$ to $128 = 4 \times 32$, respectively)} with other parameters fixed. The results are shown in Figure \ref{results of model complexity}. We can see that all models generally perform better when the number of convolutional kernels (filters) increases, and become stable after 8 kernels. We can also see that our {\ours} models outperform DQN-alpha, FCN-alpha and FCN-full with respect to the number of kernels, which indicates agent $\alpha$ can indeed learns better policies via federated learning with agent $\beta$. Both {\ours}-1 and {\ours}-2 are close to DQN-full with respect to the number of filters, suggesting that our federated learning framework is effective even though the data of agents $\alpha$ and $\beta$ are not shared with each other. 
\section{Conclusion}
we propose a novel reinforcement learning approach to considering privacies and federatively building Q-network for each agent with the help of other agents, namely Federated deep Reinforcement Learning ({\ours}). To protect the privacy of data and models, we exploit Gausian differentials on the information shared with each other when updating their local models. We demonstrate that the proposed {\ours} approach is effective in building high-quality policies for agents under the condition that training data are not shared between agents. In the future, it would be interesting to study more members in the federation with background knowledge represented by (probably incomplete) action models \cite{DBLP:journals/ai/ZhuoM014,DBLP:journals/ai/Zhuo014,DBLP:journals/ai/ZhuoK17}.

\newpage

\bibliography{FRL}
\bibliographystyle{icml2020}

\end{document}